\documentclass[10pt,twocolumn,letterpaper]{article}
\usepackage[accsupp]{axessibility}  
\usepackage{iccv}
\usepackage{times}
\usepackage{epsfig}
\usepackage{graphicx}
\usepackage{amsmath}
\usepackage{amssymb}
\usepackage{subfigure}
\usepackage{amsfonts}
\usepackage{multirow}
\usepackage{sidecap}
\usepackage{wrapfig}
\usepackage[ruled,vlined,algo2e,linesnumbered]{algorithm2e}
\usepackage{url}
\usepackage{booktabs} 
\usepackage[pagebackref=true,breaklinks=true,letterpaper=true,colorlinks,bookmarks=false]{hyperref}

\iccvfinalcopy 

\def\iccvPaperID{2143} 

\ificcvfinal\fi

\newcommand{\rev}[1]{\textcolor{black}{{#1}}}

\begin{document}

\title{GDP: Stabilized Neural Network Pruning via Gates with Differentiable Polarization}

\author{
Yi Guo\textsuperscript{1}, 
Huan Yuan,\textsuperscript{1}, 
Jianchao Tan\textsuperscript{1}, 
Zhangyang Wang\textsuperscript{2}, 
Sen Yang\textsuperscript{1}, 
Ji Liu\textsuperscript{1}\\
\textsuperscript{1}Kuaishou Technology, 
\textsuperscript{2}University of Texas at Austin
\\
\small \texttt{\{guoyi03,yuanhuan,jianchaotan,senyang,jiliu\}@kuaishou.com}, 
\small \texttt{\{atlaswang\}@utexas.edu}
}







\maketitle
\ificcvfinal\thispagestyle{empty}\fi

\begin{abstract}
Model compression techniques are recently gaining explosive attention for obtaining efficient AI models for various real time applications. 
Channel pruning is one important compression strategy, and widely used in slimming various DNNs. Previous gate-based or importance-based pruning methods aim to remove channels whose ``importance" are smallest.
However, it remains unclear what criteria the channel importance should be measured on, leading to various channel selection heuristics. 
Some other sampling-based pruning methods deploy sampling strategy to train sub-nets, which often causes the training instability and the compressed model's degraded performance.
In view of the research gaps, we present a new module named Gates with Differentiable Polarization (\textbf{GDP}), inspired by principled optimization ideas. 
GDP can be plugged before convolutional layers without bells and whistles, to control the on-and-off of each channel or whole layer block. 
During the training process, the polarization effect will drive a subset of gates to smoothly decrease to  \textbf{exact zero}, while other gates gradually stay away from zero by a large margin.
When training terminates, those zero-gated channels can be painlessly removed, while other non-zero gates can be absorbed into the succeeding convolution kernel, 
causing completely no interruption to training nor damage to the trained model. Experiments conducted over CIFAR-10 and ImageNet datasets show that the proposed GDP algorithm achieves the state-of-the-art performance on various benchmark DNNs at a broad range of pruning ratios. We also apply GDP to DeepLabV3Plus-ResNet50 on the challenging Pascal VOC segmentation task, whose test performance sees no drop (even slightly improved) with over 60\% FLOPs saving.

\end{abstract}
\section{Introduction}

Model compression techniques recently attract many attentions from both industry and academic research communities for compressing Deep Neural Networks (DNNs) to satisfy devices with limited computing and storage capabilities. 
One common model compression method is pruning, which includes unstructured pruning (i.e., weights sparsification) ~\cite{chao2020directional, han2015learning,lee2021layeradaptive,lin2020dynamic,srinivas2017training} and structured pruning (e.g., channel pruning and kernel pruning)~\cite{Yang_2018_ECCV}. Unstructured pruning removes individual weight, which results in sparse connection and requires sparse matrix operations supports. The structured pruning eliminates an entire neuron or a block at once, leaving the model's structure and connections intact and thus having no special requirements for hardware or specific libraries. In this paper, we focus on the latter.

\begin{figure}
\centering
\subfigure[Distribution of weights in LASSO]{
\begin{minipage}[t]{0.24\textwidth}
\centering
\includegraphics[width = 1.0\textwidth]{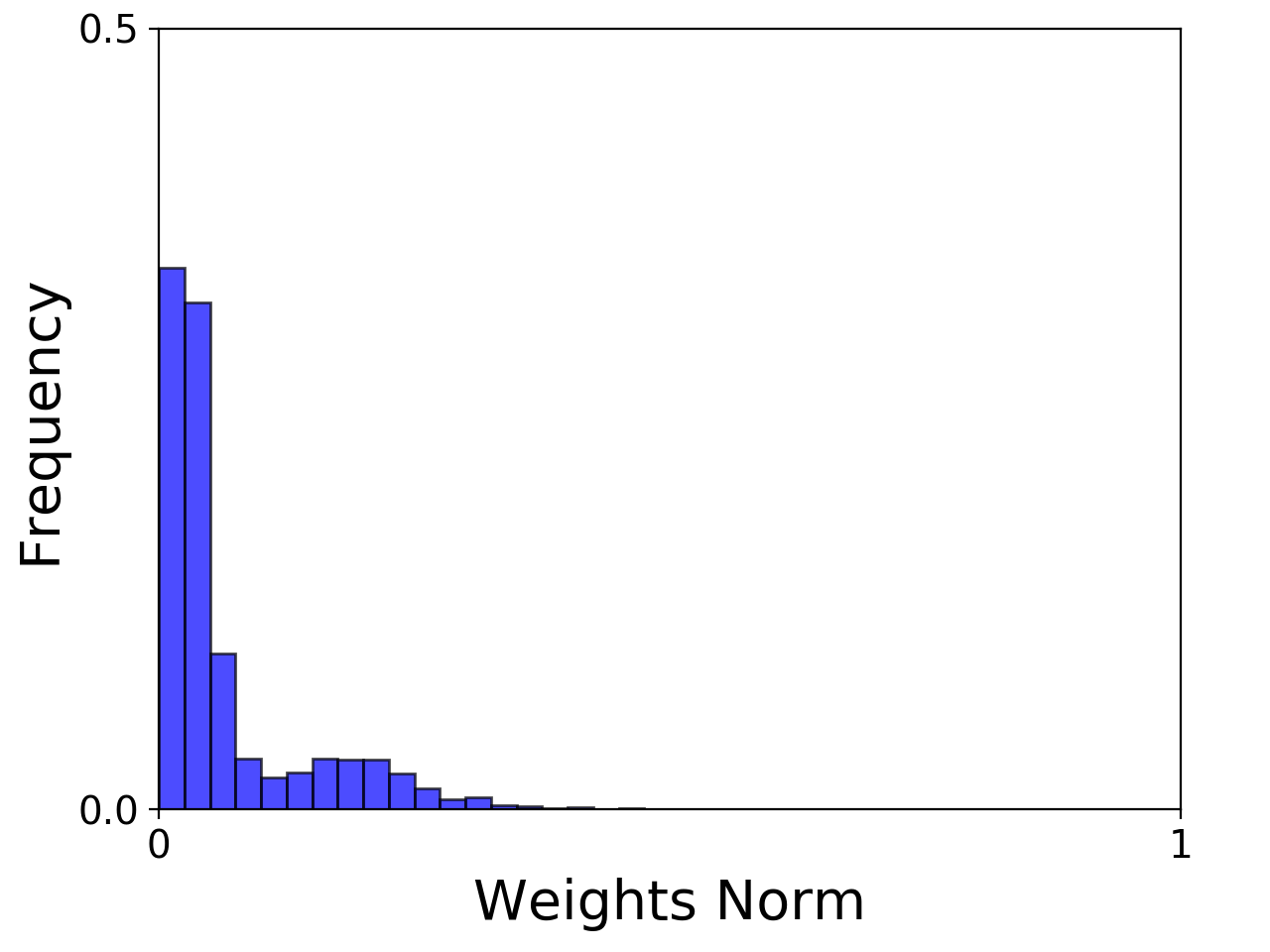}
\label{fig: lasso distribution}
\end{minipage}
}\subfigure[Distribution of gates in GDP]{
\begin{minipage}[t]{0.24\textwidth}
\centering
\includegraphics[width = 1.0\textwidth]{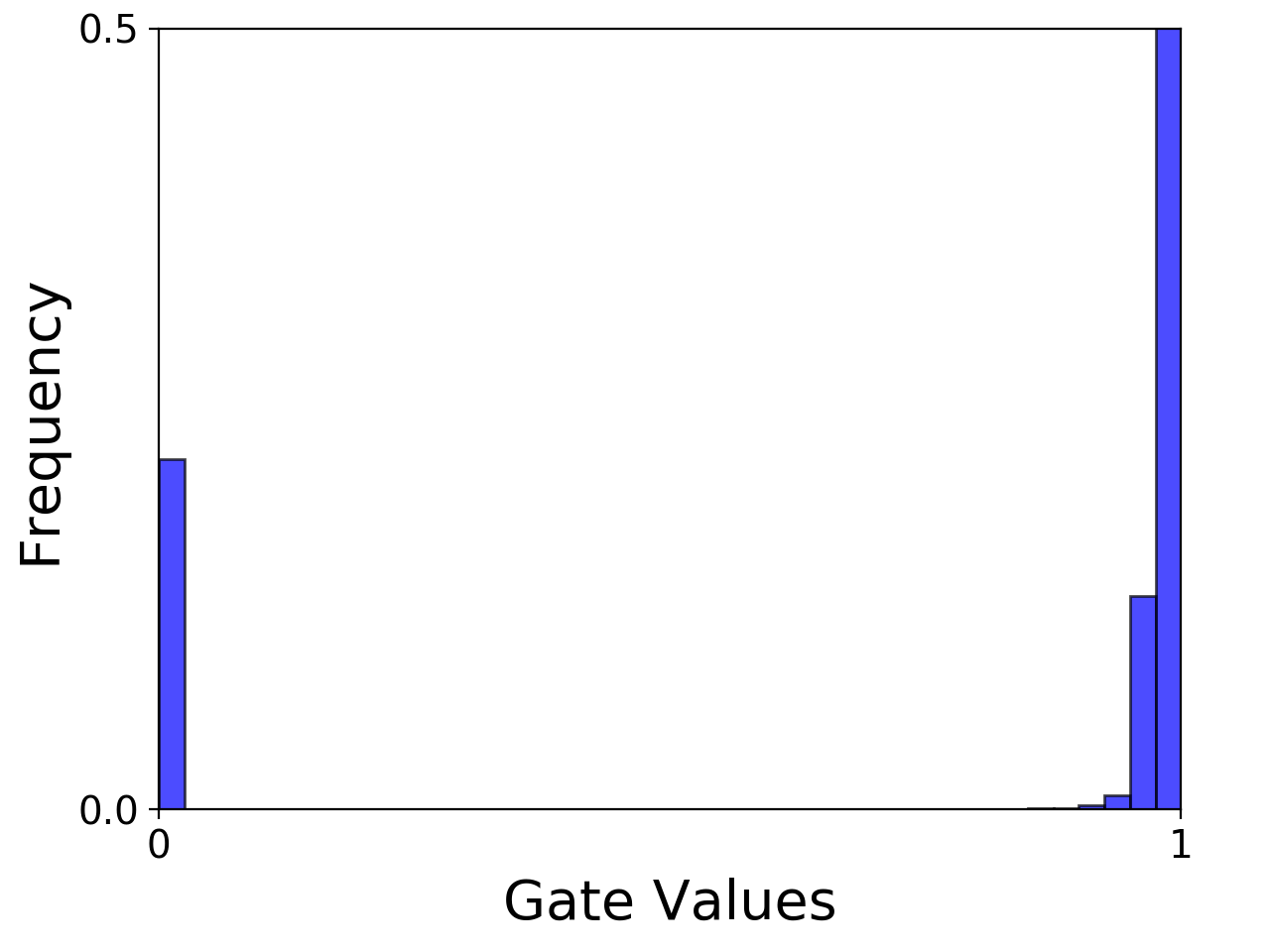}
\label{fig: sl0 distribution}
\end{minipage}
}
\caption{
The density distribution of weights norm for LASSO-based pruning (left) and gates value for GDP pruning (right). LASSO is performed on the dimension of input channel and all the norm is divided by the number of entries then scaled to $0\sim 1$. The Y-axis is truncated to 0.5 for better visualization. We can see that the LASSO method yields a continuous distribution and needs to carefully choose a small threshold to prune the neurons. While in GDP, some gates values are exact zero while the others are away from zero by a large margin. The experiment is conducted on ImageNet with MobileNet-V1.}
\label{fig: LASSO vs SL0}
\end{figure}

Importance-based pruning methods usually prune channels/blocks whose importance are small, but it remains unclear what criteria the importance should be measured on, leading to various importance selection heuristics, such as the magnitude of weights~\cite{wen2016learning}, first-order or second-order information of weights~\cite{ding2019global,molchanov2019importance,peng2019collaborative}, knockoff features of filters~\cite{tang2020scop} and so on. 
One recent track of works follow the sparse learning optimization routines, utilizing LASSO or group LASSO ~\cite{he2017channel,liu2017learning,wen2016learning} to regularize training weights.
However, these methods need to prune from a skewed yet continuous histogram of weights values, and the selection of the pruning threshold~\cite{han2015learning} is critical yet ad-hoc, taking Figure~\ref{fig: lasso distribution} for example. 
Moreover, removing many small yet non-zero weights will inevitably damage the network, and often need to fine-tune before the next pruning schedule, causing iterative pruning-finetuning loop~\cite{han2015deep, li2019oicsr,liu2017learning,luo2017thinet}. 


Besides the importance-based pruning methods, some sampling-based methods~\cite{gao2020discrete,kang2020operation,luo2020autopruner, zhao2020differentiable,he2020learning, yuan2020growing} either sample a sub-network or calculate expectations over whole network to train, utilizing Gambul-Softmax~\cite{jang2017categorical}, STE~\cite{bengio2013estimating} or \rev{Bernoulli distribution ~\cite{srinivas2017training, yuan2020growing} }strategies. 
These methods may potentially cause either unstable training due to equiprobable sampling (i.e. sample a completely different sub-net when the probability is nearly uniformly distributed at the beginning of training), or converging at local minimum due to deterministic sampling (i.e. sample nearly the same sub-net when the probability becomes deterministic), thus need various ad-hoc manual designs~\cite{gao2020discrete}.

\begin{table}
	\centering
	\footnotesize
	\begin{center}
	\setlength{\tabcolsep}{0.8mm}{
	\begin{tabular}{cccccc}
		\toprule
		FLOPs ratio & Before removing	    &   After removing  &   After fine-tuning\\
		\midrule
        43.8\% & \textbf{69.962}	&	\textbf{69.962} & 70.150 	\\
		\bottomrule
	\end{tabular}
	}
	\end{center}	
	\caption{
	This table shows there is no performance drop before and after removing zero-gated channels/blocks using GDP. This experiment is conducted on ImageNet~\cite{deng2009imagenet} with MobileNet-V2~\cite{sandler2018mobilenetv2}.}
	\label{tab:befor_after_pruned}
\end{table}

To bridge these research gaps, we introduce a novel scheme called \textit{Gates with Differentiable Polarization} (\textbf{GDP}), which can easily plug-and-play with any existing convolutional or fully connected layer. 
It does not rely on any specific module such as Batch Normalization (BN)~\cite{ioffe2015batch} or ReLU~\cite{glorot2011deep}, and contains fewer hyper-parameters to tune during optimization than those minimax optimization methods~\cite{li2019admm,yang2019ecc} used in resource-constrained model compression. 
The core idea of GDP is to encourage gate values to update smoothly towards polarization (some are \textbf{exact zero}, others stay away from zero by a large margin) during training, so that the exact-zero-gated channels/blocks can be painlessly removed and other gates be absorbed into conv/fully-connected kernels, without causing any damage, as illustrated in Figure~\ref{fig: sl0 distribution} and Table~\ref{tab:befor_after_pruned}.

Our main highlights are summarized as follows:
\begin{itemize}
\item \textbf{Core Idea of Polarization:} GDP is designed to push a large interval between \textbf{exact zero} gates and non-zero gates while preserving the gradient-friendly optimization. 
This is particularly favored, and enables us to directly remove zero-gated channels/blocks and absorb other gates to the conv/fully-connected kernels with no performance impact when training terminates.

\item \textbf{Stability, Consistency and Light-Weight:} GDP gets rid of ad-hoc thresholding nor any probabilistic sampling/expectation. That makes it (1) easy to use, with nearly no manual crafting; (2) produce stable and smooth training; 
(3) incur almost no computational or time overhead; and (4) not rely on any specific modules such as BN or ReLU.


\item  \textbf{State-of-the-Art Results:} Experiments on CIFAR-10 and ImageNet classification tasks show that GDP outperforms all previous competitors with clear margins. GDP can obtain slightly improved performance in the Pascal VOC segmentation task with DeepLabV3Plus-ResNet50, while reducing more than 60\% FLOPs.
\end{itemize}

\section{Related Work}
Model compression techniques are broadly applied in many research fields successfully, such as GAN~\cite{wang2020gan, fu2020autogan}, Recommendation system~\cite{shen2021umec} and so on. 
There are many model compression techniques proposed to solve different problems, including network pruning~\cite{iandola2016squeezenet,liu2017learning,zhu2017prune}, weights quantization~\cite{jacob2018quantization,jin2020adabits,krishnamoorthi2018quantizing}, knowledge distillation~\cite{2015Distilling,huang2017like}, architecture searching~\cite{liu2018darts,zoph2016neural}, weights hashing~\cite{eban2020structured} and so on. 
Some works~\cite{han2015deep,polino2018model,wang2020apq, yang2020automatic,gui2019model} have tried to unify aforementioned techniques into one framework, which can slim network comprehensively. In this paper, we only focus on pruning strategy to achieve SOTA compression results by leveraging a differentiable polarized gate design and we want to emphasize that our GDP method can be adjusted into a unified compression framework easily.

\subsection{Pruning Strategies}
\paragraph{Importance based pruning:}
One straightforward way to prune a network is to throw out less important components. 
~\cite{he2017channel, li2019oicsr,li2020group, liu2017learning, wen2016learning} use group LASSO to regularize convolution kernels or other scale factors during training, leaving some weights values small. However, these works suffer from how to choose pruning threshold. Furthermore, ~\cite{ye2018rethinking} points out that channels or blocks with smaller weights norm may not necessarily means that they are less important.
Besides weights norm metric, some other works ~\cite{chin2020towards,lee2018snip,molchanov2019importance,park2020lookahead,lee2021layeradaptive,lubana2021a} explore some new importance metrics, such as first-order, second-order taylor expansions of weights and so on. The designing of importance metrics is still an open problem to explore. 

\paragraph{Sampling based pruning:}
Sampling process is not differentiable, however ~\cite{he2020learning,kang2020operation} use Gumbel-Softmax~\cite{jang2017categorical} trick to make sampling process differentiable during network training, thus can be used in pruning network. While~\cite{gao2020discrete} uses straight-through estimator~\cite{bengio2013estimating} on a step function controlled by probability oscillating around 0.5. 
\rev{~\cite{louizos2018learning, yuan2020growing}} relaxes the intractable L0 regularization by using hard-sigmoid function as gate function instead, then sample a sub-net by Gumbel-Softmax similarly as previous work or \rev{by Bernoulli ~\cite{yuan2020growing}}. 
\rev{These methods sample sub-networks stochastically, while we continually adapt a single network.}
DARTS~\cite{liu2018darts} introduced differentiable architecture search by formulating the output of a layer as the expectation of all input candidate edges and then choosing edges with the largest probabilities, which can be treated as a general pruning method. We do not want to compute expectation, since it cannot accurately depict real sub-net during training.  

\paragraph{Other variants:}
Beyond above two categories, several works choose to prune networks gradually by reinforcement learning~\cite{he2018amc, tan2019mnasnet}; meta-learning~\cite{liu2019metapruning}; evolutionary methods~\cite{guo2020dmcp} and so on.
Furthermore, ~\cite{yu2019autoslim} trains a proxy network to represent the accuracy of sub-nets with different channel configurations, and then greedily slim a network.
~\cite{Cai2020Once-for-All:} considers a generalized pruning on all dimensional of network to progressively train once for all sub-nets.
All these methods may suffer from the computational resources overhead and their complicated designs may hinder the popularization and application of them.
However, our pruning method is straightforward and takes about the same amount of time as training of baseline model. 

A series of recent works ~\cite{li2019admm,ren2019admm,yang2019ecc,zhang2018systematic} formulate the network pruning problem into a constrained minimax optimization problem, and then use ADMM method~\cite{boyd2011distributed} to solve it. However, these methods inevitably introduce many hyper-parameters (like dual variables) to tune. Instead, our method is simple yet efficient, with fewer parameters to tune during compression.

\section{Methods}
We introduce the details of our pruning algorithm in this section, including a special gate design, the FLOPs reformulation based on it and the proposed alternative proximal optimization method.
\subsection{Pruning via Gates}
\begin{figure}
\centering
\includegraphics[width = 0.42\textwidth]{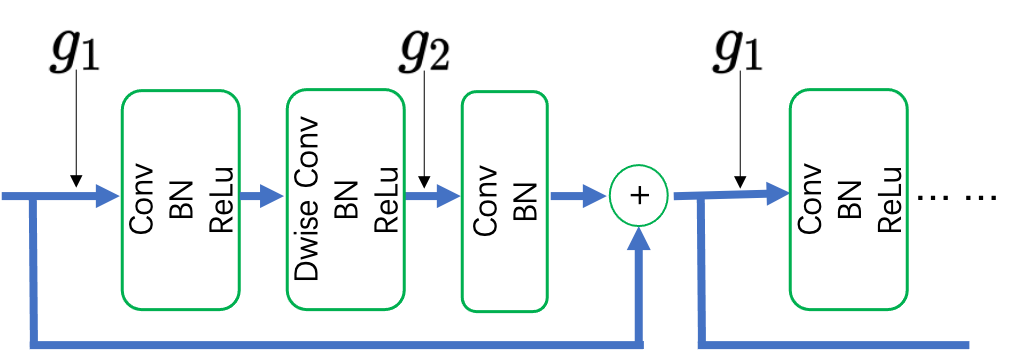}
\caption{Take a residual-block as example to show the position where the gates plug into. The gates are plugged immediately before regular convolution layers but not depth-wise convolution. 
Note that $g_1$ appears twice to ensure consistency of the two convolution kernels connected by skip connection. If all the entries of $g_2$ become zeros, the entire block can be removed (except for the bias term in the BN).}
\label{fig: resnet}
\end{figure}

For CNNs compression, instead of directly measuring the channel importance as in previous work, we plug trainable gates 
immediately before regular convolution (`regular' is used to distinguish from `depth-wise') or fully connected layers to control the on-and-off of each channel or even whole block. Some works~\cite{kang2020operation,ye2018rethinking, you2019gate} utilize BN or ReLU layer as gates, however, a channel being judged as unimportant 
now doesn't mean it won't be important after a series of channel-wise operations, such as pooling, depth-wise convolution~\cite{chollet2017xception}, \etal. What's more, some networks may not contains BN or ReLU. So, we plug gates immediately before the position where information from different channels are going to merge. Different from ~\cite{gao2020discrete}, we do not plug gates before depth-wise convolution, which does not merge information from different channels. Figure~\ref{fig: resnet} illustrates the positions where the gates should be plugged into, without losing generalization, simply taking a residual block in MobileNet-V2 as example.

Specifically, let $X_{l-1} \in \mathbb{R}^{c\times h \times w}$ be the input of the $l$-th layer convolution kernel $W_l \in \mathbb{R}^{c\times d \times r \times r}$, where $c$, $h$ and $w$ represent the number of channels, height and width of the input feature map respectively; $d$ and $r$ is the number of output channels and kernel size of the convolution kernel, respectively. 
We can insert a gate vector $\bar{g_l}=\{g^{(1)}_l, g^{(2)}_l,...,g^{(c)}_l\}$ with dimension $c$ as channel-wise scaling factors for $X$, before it is convolved by $W$. 
That is, the output, $Y_l \in \mathbb{R}^{d\times h \times w}$, of the convolution operation is 
\begin{equation}
    Y_l = W_l * (\bar{g_l} \odot X_{l-1})
\end{equation}
where $\odot$ is the channel-wise multiplication, and $*$ is the convolution operation. One can easily control the on-and-off of these channels by manipulating $\bar{g_l}$. In GDP, some gates smoothly decrease to be exact zeros, while others stay away from zeros by a large margin gradually. Then we can safely remove channels corresponding to the zero-gates, and absorb the others to the successive convolution kernel. By this way, the sub-net can be got with performance same to the super-net.

\subsection{Differentiable Polarized Gates}
Previous gate-based methods usually use some optimization tricks to make the gate differentiable, such as Gumbel-Softmax trick used in sampling-based method and L1-norm penalty in regularization based method and so on. These methods either cause unstable training process or cannot obtain exact zero values. To ensure the stability and effectiveness of model pruning, the desired gate should have following properties: 1. be differentiable for back-propagation with simpler tricks than complicated sampling mechanism; 2. the values are better to range in $[0,1]$; 3. the distribution of values should be two modes, one mode keeps exact zeros while others stay away from zeros obviously. 
Also, we do not want to construct an expectation-based or sampling-based resource calculation, which can not represent the accurate resource of a specific sub-net. \rev{~\cite{srinivas2017training} constructs a heuristic bi-modal regularizer to satisfy above conditions, while we choose the smoothed L0 formulation from principled optimization methods~\cite{xiang2019new}:}
\begin{equation}
    g_{\epsilon}(x)=\frac{x^2}{x^2+\epsilon}
    \label{eq:smoothed l0}
\end{equation}
where $\epsilon$ is a small positive number. This function has the following important property:

\begin{equation}
    g_{\epsilon}(x)
    \begin{cases}
    =0,\quad x=0\\
    \approx 1, \quad x \ne 0
    \end{cases}
    \label{eq:property}
\end{equation}

\begin{figure}
\centering
\subfigure[$g_{\epsilon}$ with different $\epsilon$]{
\begin{minipage}[t]{0.24\textwidth}
\centering
\includegraphics[width = 1.0\textwidth]{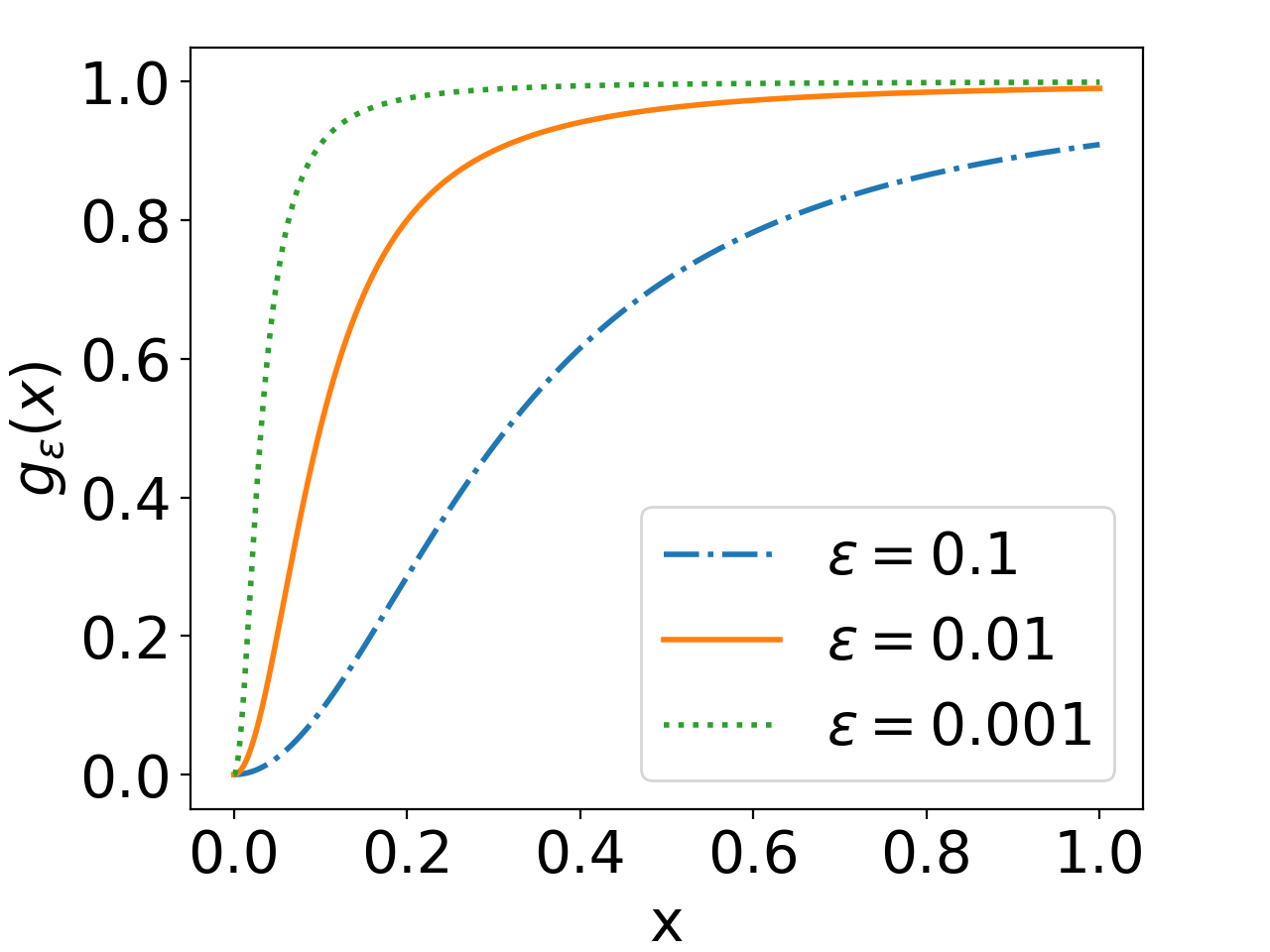}
\end{minipage}
}\subfigure[$g'_{\epsilon}$ with different $\epsilon$]{
\begin{minipage}[t]{0.24\textwidth}
\centering
\includegraphics[width = 1.0\textwidth]{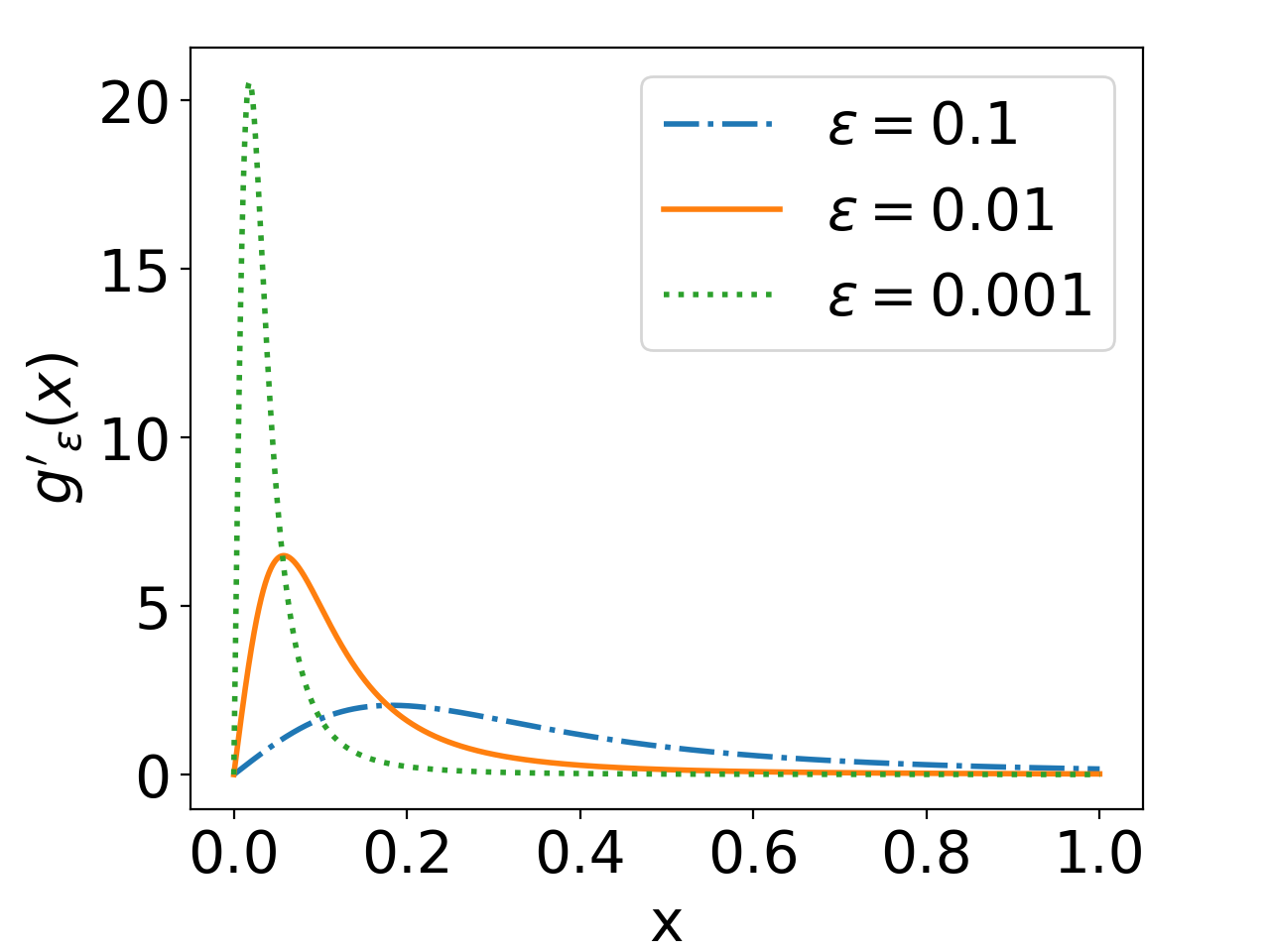}
\end{minipage}
}
\caption{The graph of $g_{\epsilon}$ (left) and $g'_{\epsilon}$ (right). As the $\epsilon$ approaching to zero, $g_{\epsilon}$  becomes polarized, and $g'_{\epsilon}(x)$ is zero only at the points where $g_{\epsilon}(x)$ get exact zeros or near ones, otherwise infinity.}
\label{fig: g and gd}
\end{figure}

The derivative of $g_{\epsilon}(x)$ w.r.t $x$ is 
\begin{equation}
    g'_{\epsilon}(x)=\frac{2x\epsilon}{(x^2+\epsilon)^2}
\end{equation}
The graph of $g_{\epsilon}(x)$ and $g'_{\epsilon}(x)$ are shown in Figure \ref{fig: g and gd}.

We can observe that 
when $\epsilon$ is small enough,  $g'_{\epsilon}(x)$ approaches to zero only at the points where $g_{\epsilon}(x)$ get exact zeros or ones, otherwise infinity. It means that the function can only stay stable at 0 or 1. Different from Gumbel-Softmax with decreasing temperature, Function~\eqref{eq:smoothed l0} can be exactly zero even when $\epsilon$ is not small. This good quality implies to us that we can regard $g_{\epsilon}(x)$ as gate, plugging into the neural network we want to compress. The subscript $\epsilon$ will be omitted in the following text when not causing misunderstanding. Theoretically, $x$ here can either be an newly introduced trainable parameters, or can directly be the norm of convolution kernel weights. We will analyze both settings in ablation study. We choose the former setting in our algorithm, since it works better than latter one. 

During training, $\epsilon$ gradually decays by a factor smaller than one, and all the gates $g^{(i)}, i=1,2,...,c$ vary smoothly between 0 and 1. When the training terminates, $\epsilon$ becomes small enough, and some gates become exact zeros while the others approach ones. Then, we can prune network explicitly by removing the channels corresponding to zero-gates and absorbing other close-to-ones gates into convolution kernel $W$. In almost all of our experiments, the performance of the pruned-net is exactly same as that of the super-net before explicitly pruning. In section~\ref{experiments}, we will show that the explicitly pruned sub-nets without fine-tuning,  already keep pace with or even better than previous state-of-the-art works. We think that the performance consistency between pruned-net and super-net is a good property to help us choose good candidate sub-nets.

\subsection{Resource-constrained Pruning Formulation}
Some works~\cite{gao2020discrete,zhao2020differentiable} calculate the resource by expectation, which do not represent the accurate resource of a specific sub-net. Instead, we formulate the resource by a bilinear function with respect to the number of channels of each layer~\cite{yang2019ecc}. Specifically, if the resource is FLOPs, a bilinear function can strictly model it for most common CNNs. In this paper, we directly take FLOPs as resource to illustrate our pruning algorithm. 
The FLOPs of a CNN can be accurately formulated as:
\begin{equation}
    \mathcal{R}(c)=\sum_{l,k, l \ne k}a_{lk}c_lc_k + \sum_lb_lc_l
    \label{eq:resource}
\end{equation}
Where $c_l$ is the number of channels in the $l$-th layer, $a_{lk}, b_l \in [0, +\infty)$ is the coefficients of bilinear function. 
Specifically, for the $l$-th layer with $c_l$ channels, gates $\bar{g_l}(\alpha_l) = \{g_l^{(1)}(\alpha_l^{(1)}),..., g_l^{(c_l)}(\alpha_l^{(c_l)})\}$ with introduced learnable parameters $\bar{\alpha}_l=\{\alpha_l^{(1)}, ..., \alpha_l^{(c_l)}\}$ are inserted immediately before the $l$-th convolution layer. Then, according to Function \eqref{eq:property}, 
the resource function \eqref{eq:resource} can be rewritten as:
\begin{equation}
    \mathcal{R}(\alpha)=\sum_{l,k, l \ne k}a_{lk}\|\bar{\alpha}_l\|_{0}{\Vert \bar{\alpha}_k \Vert}_0 + \sum_lb_l{\Vert \bar{\alpha}_l \Vert}_0
    \label{eq:resource-rewritten}
\end{equation}
where ${\Vert \cdot \Vert}_0$ is $\ell_0$-norm of a vector.
The final objective function can be formulated as below:
\begin{equation}
    \min_{w,\alpha}\mathcal{F}(w;\alpha) = \mathcal{L}(w;g(\alpha))+\lambda \mathcal{R}(\alpha)
    \label{eq:objective}
\end{equation}
where $\mathcal{L}$ is the original training loss function, $w$ denotes all the learnable weights of original network, $\alpha = \{\bar{\alpha}_1, ..., \bar{\alpha}_L\}$ denotes all the introduced learnable parameters for gates, $L$ is the total number of layers, and $\lambda$ is a balance factor.

\subsection{Optimization via Alternative Relaxation}
Objective function \eqref{eq:objective} is not easy to solve due to the existence of $\ell_0$-norm. $w$ and $\alpha$ in $\mathcal{L}(w;g(\alpha))$ can be updated by common gradient descent algorithm, just as did in normal training of baseline model. For $\mathcal{R}(\alpha)$, we can use proximal-SGD~\cite{nitanda2014stochastic} to update $\alpha$. The proximal operator $ prox_{\eta_1 \lambda \mathcal{R}(\alpha)}(\hat{\alpha})$ is defined as:
\begin{equation}
\mathop{\arg\min}_{\alpha} \frac{1}{2} \| {\alpha}
-\hat{\alpha}\|^2 + \eta_1 \lambda \mathcal{R}(\alpha)
\label{eq:proximal}
\end{equation}
Formulation \eqref{eq:proximal} is a proximal problem with a bi-linear $\ell_0$-norm format. We solve it iteratively by alternative relaxation. Specifically, we alternately update $\bar{\alpha}_l$ by relaxing $\|\bar{\alpha}_l\|_0$ to $\|\bar{\alpha}_l\|_1$ while treating other $\bar{\alpha}_k, k \ne l$ as constants at current time, and then exchange the variables and constants roles to perform relaxing operation again. The details are shown in Algorithm~\ref{alg1}. Note that in line~\ref{line4}, we do not relax $\|\bar{\alpha}_k\|_0$ to $\|\bar{\alpha}_k\|_1$ when it is treated as constant.

\begin{algorithm2e}[htbp]
\small
\label{alg1}
 \SetAlgoLined
 \KwIn{learning rates $\eta_1$, balance factor $\lambda$, predefined iteration number $N$, and $\hat{\alpha}$.}
 \KwResult{$\alpha^*$}
 Initialize $t=1, \alpha^1$ (initial values); \\
	\While{$t\leq N$}{
	\For{$l=1 \to L$}{
    $\alpha_l^{t+1} =  \mathop{\arg\min}_{\bar{\alpha}_l} \frac{1}{2} \| {\bar{\alpha}_l}
-\hat{\bar{\alpha}}_l\|^2 + \eta_1 \lambda \sum_{k, k \ne l}a_{lk}\|\bar{\alpha}_k^t\|_{0}{\Vert \bar{\alpha}_l \Vert}_1 + b_l{\Vert \bar{\alpha}_l \Vert}_1$
\label{line4}\\
	}
	$t=t+1$
}
$\alpha^* = \alpha^{N}$
\caption{Alternative relaxation to solve~\eqref{eq:proximal} iteratively.}
\end{algorithm2e}

Line \ref{line4} in Algorithm~\ref{alg1} is proximal $\ell_1$-norm problem, which has closed form solution:
\begin{equation}
\alpha_{l, i}^{t+1} =
    \begin{cases}
    \hat{\alpha}_{l,i}-\beta, \quad \hat{\alpha}_{l,i} \ge \beta \\
    0, \quad -\beta < \hat{\alpha_l}_{,i} < \beta \\ 
    \hat{\alpha}_{l,i}+\beta, \quad \hat{\alpha}_{l,i} \le -\beta \\
    \end{cases}
\label{eq: closed-form prox-l1}
\end{equation}
where $\beta=\eta_1 \lambda \sum_{k, k \ne l}a_{lk}\|\bar{\alpha}_k^t\|_{0} + b_l$, and the subscript $i$ means the $i$-th entry of the vector.

\begin{figure}[t]
  \includegraphics[width = 0.24\textwidth]{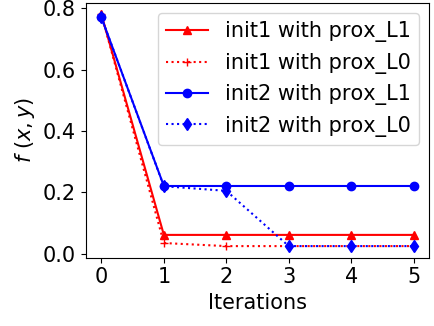}
  
  \vspace*{\dimexpr-\parskip-8\baselineskip}
  \parshape 10 
    .26\textwidth 0.23\textwidth 
    .26\textwidth 0.23\textwidth 
    .26\textwidth 0.23\textwidth
    .26\textwidth 0.23\textwidth
    .26\textwidth 0.23\textwidth
    .26\textwidth 0.23\textwidth
    .26\textwidth 0.23\textwidth
    .26\textwidth 0.23\textwidth
    .26\textwidth 0.23\textwidth
    0pt 0.5\textwidth 
  \makeatletter
  \refstepcounter\@captype
  \addcontentsline{\csname ext@\@captype\endcsname}{\@captype}
    {\protect\numberline{\csname the\@captype\endcsname}{ToC entry}}%
  \csname fnum@\@captype\endcsname: 
  \makeatother
  \small{A simple example to show the convergence of Algorithm~\ref{alg1}. The objective function here is set to be $f(x,y)=\frac{1}{2}((x-u)^2+(y-v)^2)+c\|x\|_0\|y\|_0$, where $u=\{0.1,0.2\}$, $v=\{-0.3,0.5,0.6\}$ and $c=0.01$. init1 denotes the initialization with $x=\{0.1,0.8\}$ and $y=\{0.7,0.3,0.8\}$, while init2 denotes the initialization with $x=\{0.4, 0.1\}$ and $y=\{0.7, 0.5, 0.0\}$.}
  \label{fig: L0L1 iter}
\end{figure}

In fact, general proximal $\ell_0$-norm problem also has closed form solution. 
The reason we choose to relax $\ell_0$-norm to $\ell_1$-norm in our algorithm is that all entries of $\alpha$ in proximal $\ell_1$-norm operation can shrink a small step at each iteration, which may compete with the change caused by $\mathcal{L}(w;g(\alpha))$. 
This effect exists regardless of the initialization value of $\alpha$.
However, for proximal $\ell_0$-norm operation, the entries with large initialization values will always fluctuate in a small range, while the other entries initialized with small values quickly vanish at the first few iterations and very hard to recover later by $\mathcal{L}(w;g(\alpha))$. 
We find in our experiments that proximal $\ell_1$-norm operation works robustly while proximal $\ell_0$-norm operation is very sensitive to initialization of $\alpha$.\\
As for convergence of Algorithm~\ref{alg1}, with large number of simulations, we find that it always converges quickly, although it's not necessarily the globally optimal solution. An simple illustration example is shown in Figure \ref{fig: L0L1 iter}, which converge quickly with very few iterations. In all experiments in this paper, we find iteration $N=1$ works well enough.

\section{Experiments}
\label{experiments}
We conduct classification experiments on CIFAR-10~\cite{krizhevsky2009learning} dataset with ResNet-56, VGG-16 and MobileNet-V2, and ImageNet ILSVRC-12 dataset~\cite{deng2009imagenet} with  MobileNet-V1/V2, ResNet-50 respectively. Our main experiments are done by introducing learnable parameters $\alpha$ while we also do ablation study with using convolution kernel as parameters for gate function in Equation \eqref{eq:smoothed l0}. 
We compare GDP with previous works: EagleE~\cite{li2020eagleeye}, PFS~\cite{wang2020pruning}, MetaP~\cite{liu2019metapruning}, AMC~\cite{he2018amc}, NetAdapt~\cite{yang2018netadapt}, DCP~\cite{zhuang2018discrimination}, GFS~\cite{ye2020good}, LeGR~\cite{chin2020towards}, DMCP~\cite{guo2020dmcp}, SCP~\cite{kang2020operation}, SLRTD~\cite{phan2020stable}, OICSR~\cite{li2019oicsr}, Taylor~\cite{molchanov2019importance}, CCP~\cite{peng2019collaborative}, GD~\cite{you2019gate}, Hinge~\cite{li2020group}, CURL~\cite{luo2020neural}, HRank~\cite{lin2020hrank}, LFPC~\cite{he2020learning}, C-SGD~\cite{ding2019centripetal}, SCOP~\cite{tang2020scop}, AutoSlim~\cite{yu2019autoslim}, ABCpruner~\cite{lin2020channel}, DMC~\cite{gao2020discrete}, VCNNP~\cite{zhao2019variational}. It should be pointed out that AutoSlim~\cite{yu2019autoslim} uses totally different training settings from most works, and prunes from more heavier models, thus not comparable. 
Also we should mention that MDP~\cite{guo2020multi} is more likely NAS method for searching sub-nets; EagleE~\cite{li2020eagleeye} uses adaptive BN strategy to improve various sub-nets evaluation. 
We did not compare with these special papers on CIFAR-10, due to their extremely good performance reported, with a large gap to other papers. 
Although aforementioned two papers are not fair to compare with our pruning method, our simple yet effective method still achieves better performance than these out-of-scope papers on ImageNet with various benchmark networks. 
The drawing scripts containing all numerical values of all figures are left in supplementary materials.
\subsection{ Implementation Details}

\paragraph{$\alpha$ and $\epsilon$ settings} For all experiments, the $\alpha$ and $\epsilon$ are initialized with 1.0 and 0.1 respectively. The learning rate of $\alpha$ is one tenth of that of the other network parameters for classification task and there is no weight decay on $\alpha$ when training. For CIFAR-10, we train 350 epochs with decaying $\epsilon$ by 0.96 every epoch, while for ImageNet, we train 140 epochs with decaying $\epsilon$ by 0.9 every epoch. For DeepLabV3Plus-ResNet50, we train it for 60k iterations, and decay $\epsilon$ by 0.97 every 120 iterations. Also, the initial learning rate of $\alpha$ is 9e-4 as it is different for classifier and backbone. We smooth the shrinkage of $\alpha$ when necessary.

\paragraph{CIFAR-10 Experiments} We randomly flipped horizontally, randomly cropped size of 32 with padding 4 on images, and perform normalization to augment the data when training. Table~\ref{tab: cifar10} shows the performance of CIFAR-10 on various network structure. The network structure for VGG-16 and ResNet-56 is same as \cite{lin2020hrank}. MobileNet-V2 is from LeGR~\cite{chin2020towards}, but we do not find the CIFAR-10 results in LeGR. Note that the structure of MobileNet-V2 for CIFAR-10 may differ in different works. We train it with cosine learning rate~\cite{loshchilov2016sgdr} without label smoothing~\cite{szegedy2016rethinking}. These experiments are conducted on single NVIDIA 2080Ti GPU.


\begin{table}[]
\small
\centering
\setlength{\abovecaptionskip}{5pt}
\setlength{\belowcaptionskip}{12pt}
\setlength{\tabcolsep}{1.4mm}{
\begin{tabular}{cccccc}
\toprule
                              & Method                & Baseline(\%) & Ratio(\%) & \rev{Abs. Acc.} & $\Delta$ Acc. \\
\hline
\multirow{6}{*}{\rotatebox{90}{VGG-16}}          & PFS   & 93.44        & 50        & 93.63 & \textbf{0.19}     \\
                              & SCP   & 93.85        & 33.77     & 93.79 & -0.06      \\
                              & VCNNP & 93.25        & 39.1      & 93.18 & -0.07      \\
                              & Hinge & \textbf{94.02}        & 60.93     & 93.59 & -0.43      \\
                              & HRank & 93.96         & 46.4      & 93.43 & -0.53      \\
                              & GDP (Ours)            & 93.89        & \textbf{30.55}     & \textbf{93.99} & 0.1      \\
\hline
\multirow{8}{*}{\rotatebox{90}{ResNet-56}}    & PFS                   & 93.23        & 50        & 93.05 & -0.18      \\
                              & ABCPruner             & 93.26        & 45.87     & 93.23 & -0.03      \\
                              & HRank                 & 93.26        & 50        & 93.17 & -0.09      \\
                              & SCOP                  & 93.7         & 44        & 93.64 & -0.06      \\
                              & LFPC                  & 93.59        & 52.9      & 93.72 & \textbf{0.13}     \\
                              & LeGR                  & \textbf{93.9}         & 47        & 93.7  & -0.2       \\
                              & GDP (Ours)      & \textbf{93.9}         & \textbf{34.36}     & 93.55 & -0.35      \\
                              & GDP (Ours)       & \textbf{93.9}         & 46.65     & \textbf{93.97} & 0.07     \\
\hline
\multirow{4}{*}{\rotatebox{90}{Mobile-V2}} & SCOP                  & 94.48        & 59.7      & 94.24 & -0.24      \\
                              & MDP                   & 95.02        & 71.29     & 95.14 & 0.12     \\
                              & DMC                   & 94.23        & 60        & 94.49 & \textbf{0.26}     \\
                              & GDP (Ours)                  & \textbf{94.89}        & \textbf{53.78}     & \textbf{95.15} & \textbf{0.26}    \\
\bottomrule
\end{tabular}}
\vspace{5pt}
\caption{CIFAR-10. \rev{Different methods' accuracy  (Abs. Acc.) and their differences from corresponding baselines ($\Delta$ Acc.) are reported for a full comparison.} Note that the structure of MobileNet-V2 for CIFAR-10 may differ in different works.}

\label{tab: cifar10}
\end{table}


\paragraph{ImageNet Experiments} We conduct experiments on ImageNet ILSVRC-12 dataset~\cite{krizhevsky2012imagenet} with MobileNet-V1/V2~\cite{howard2017mobilenets,sandler2018mobilenetv2} and ResNet-50~\cite{he2016deep}. We use this dataset the same way as PyTorch official example\footnote{https://github.com/pytorch/examples/blob/master/imagenet/main.py}. 

\begin{figure}
\centering
\begin{minipage}[t]{0.45\textwidth}
\centering
\includegraphics[width = 1.0\textwidth]{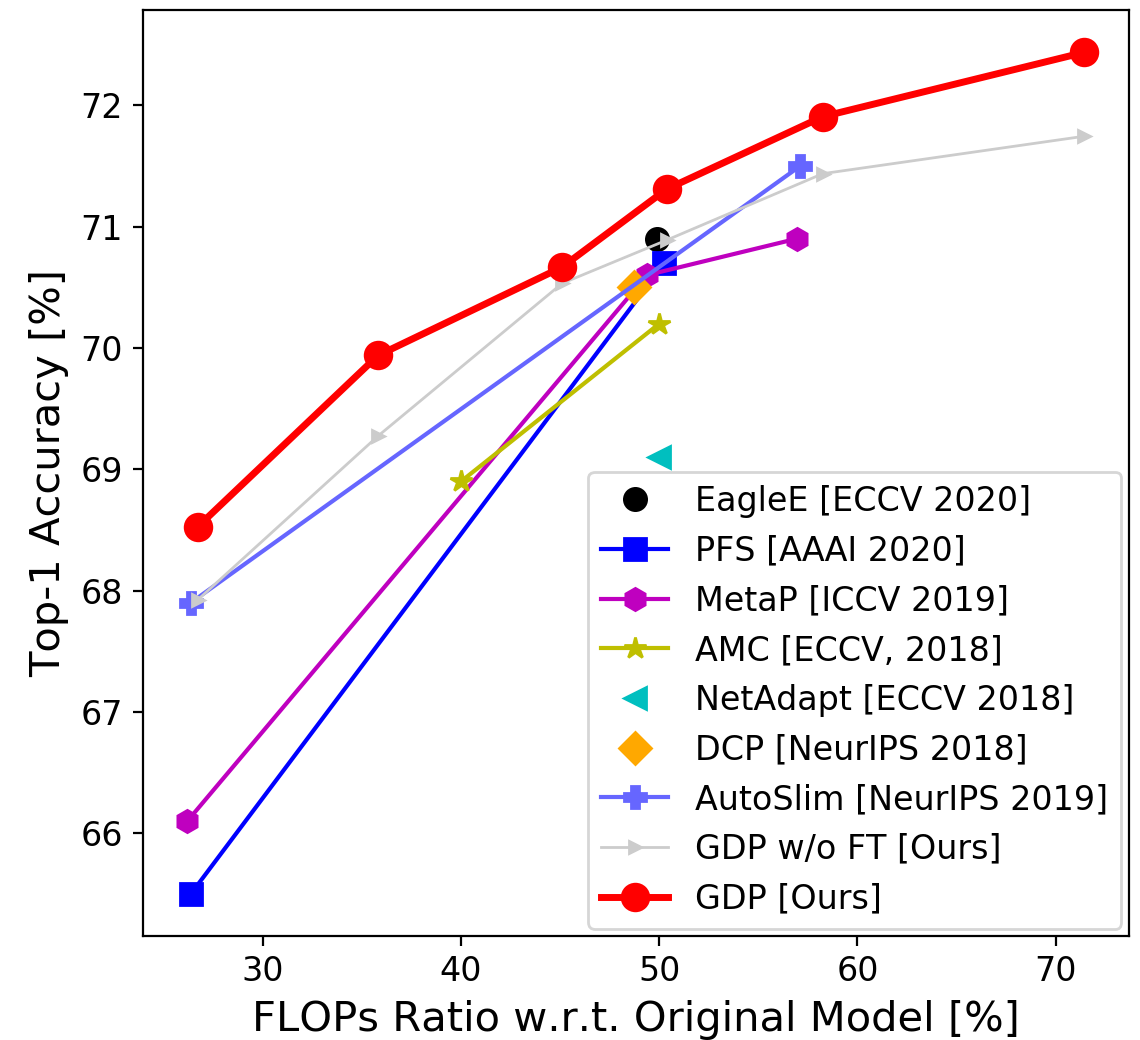}
\end{minipage}
\caption{Performance of GDP compared with other works with Top-1 Accuracy of ImageNet on MobileNet-V1. We can see the great advantages of GDP (even without fine-tuning), especially at low FLOPs ratio. Our baseline is 71.314. Note that our results are even better than AutoSlim~\cite{yu2019autoslim}.}
\label{fig: mobiV1}
\end{figure}

For MobileNet-V1, the initial learning rate is set to 0.1 and decayed by 0.1 every 35 epochs for training. No label smoothing is used, same to most previous works. Figure \ref{fig: mobiV1} shows the performance of GDP compared to other works.

\begin{figure}
\centering
\includegraphics[width = 0.450\textwidth]{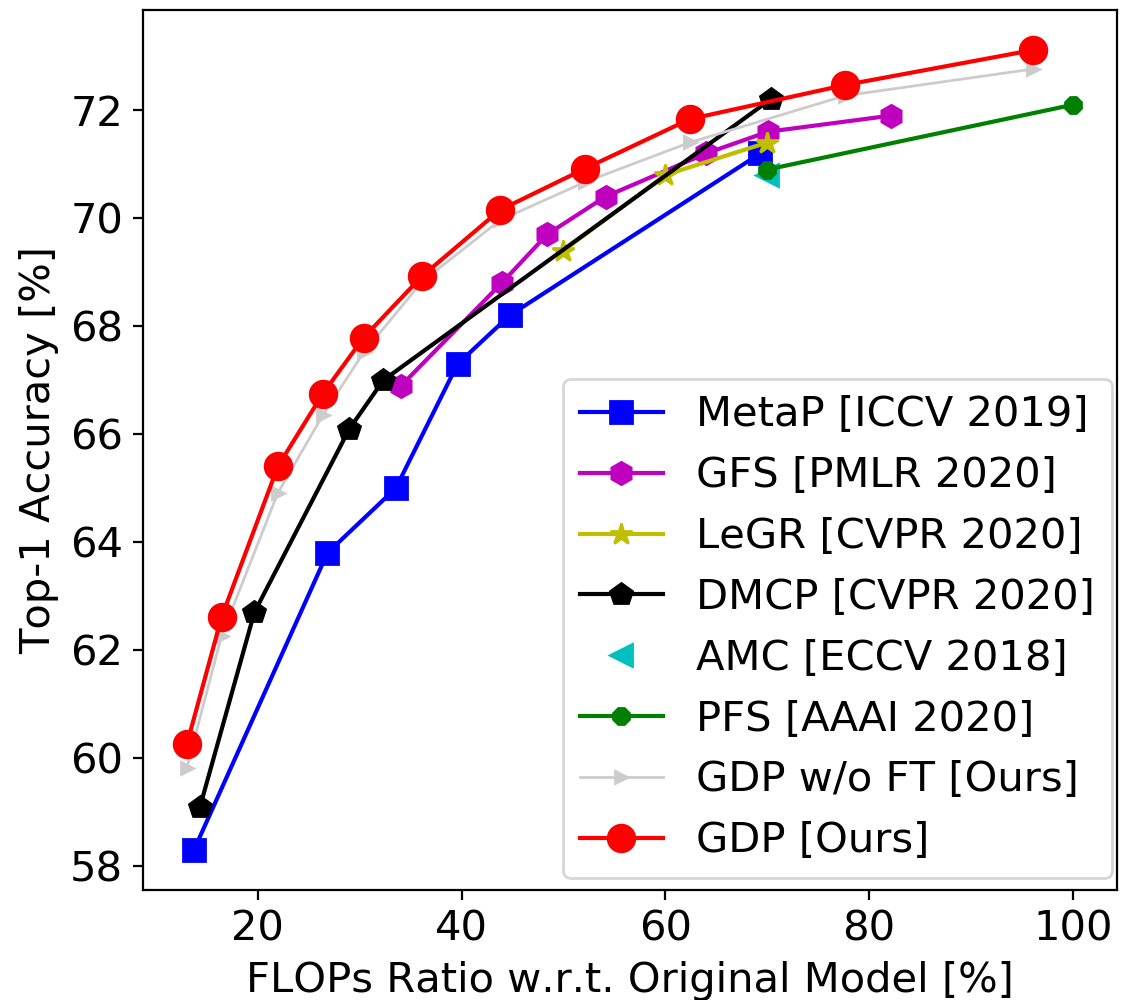}
\caption{Performance of GDP compared with other works with Top-1 accuracy of ImageNet on MobileNet-V2. Our baseline is 72.13, similar to most previous works. We can see the great advantages of GDP (even without fine-tuning), especially at the same FLOPs ratios less than 40\%, the gap of top-1 accuracy is close to two percentages \rev{(it is better to compare dots strictly vertically)}.}
\label{fig: mobiV2}
\end{figure}

MobileNet-V2 is designed with inverted residual blocks, and it is more efficient than MobileNet-V1, which is more challenging for model pruning. Same as most previous works, we use label smoothing and cosine learning rate to boost performance. Figure \ref{fig: mobiV2} shows the performance of GDP compared with other works. We can clearly see that GDP has enjoyed great advantages over previous works. We do not show AutoSlim~\cite{yu2019autoslim} here because it prune MobileNet-V2 from a more heavier super-net (MobileNet-V2 1.5$\times$) and used many training tricks. We do not use this settings to keep comparable with most others works.

\begin{figure}
\centering
\includegraphics[width = 0.450\textwidth]{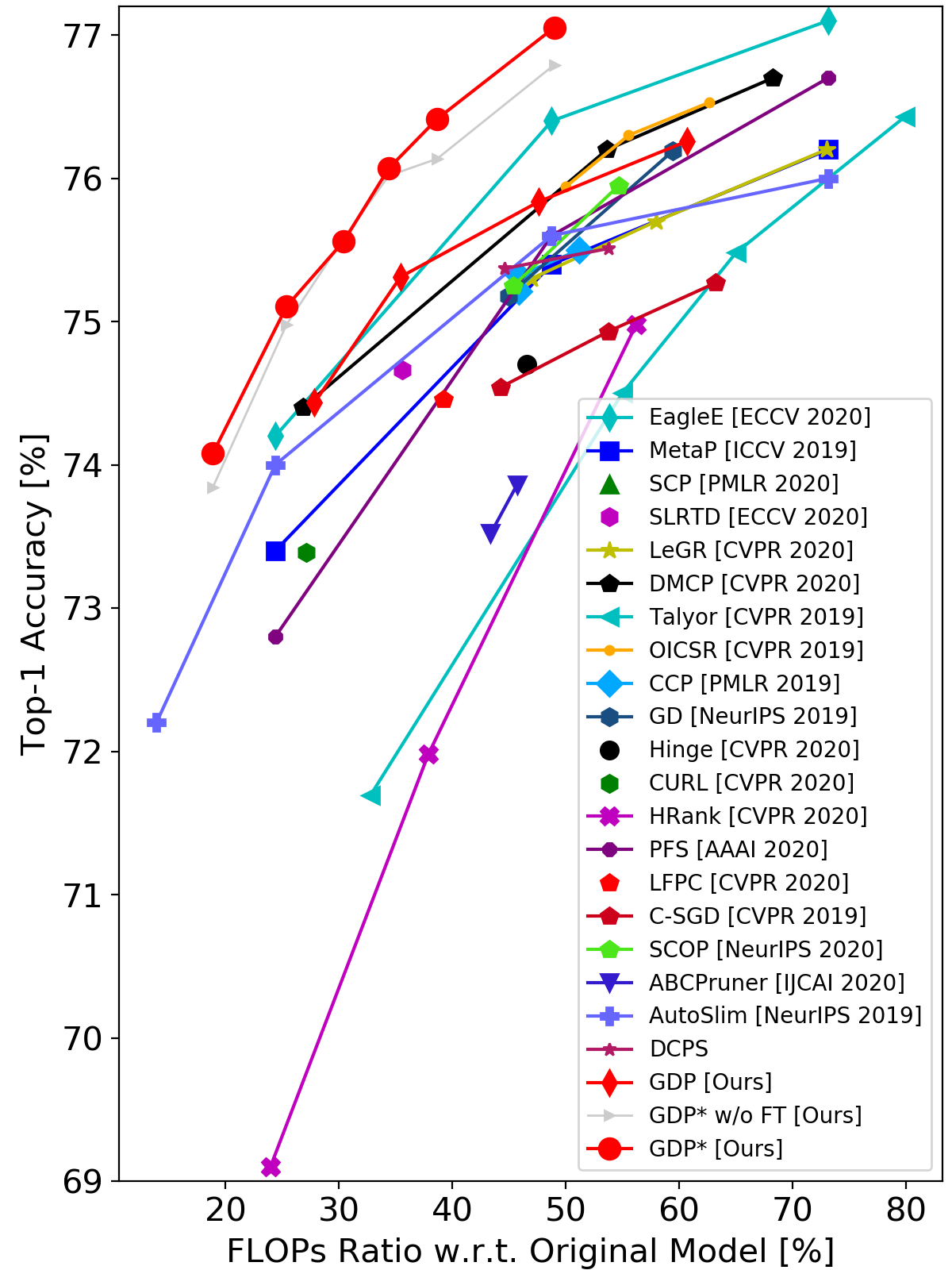}
\caption{Top-1 Accuracy of Imagenet on ResNet-50. Different training settings are used for different works, and the baseline also varies greatly. We provide two version curves here. GDP* means training with both cosine learning rate and label smoothing (baseline 77.414), while GDP use step learning rate without label smoothing (baseline 76.140).}
\label{fig: res50}
\end{figure}

ResNet-50 is a network with much more FLOPs. Some works use cosine learning rate and/or label smoothing while others not, the baseline also varies dramatically. So we plot two version curves, one with cosine learning rate and label smoothing, the other neither. See Figure~\ref{fig: res50}, GDP is also superior than other works. Experiments for MobileNet-V1/V2 are conducted on two NVIDIA 2080Ti GPUs, while experiments for ResNet-50 are on two NVIDIA V100 GPUs.


\paragraph{Pascal VOC with DeepLabv3+} DeepLabv3+~\cite{chen2018encoder} is designed with a decoder module besides the Atrous Spatial Pyramid Pooling(ASPP) to extend DeepLabv3~\cite{2017Rethinking} and get better segmentation results. It contains a more complex network structure than classification tasks, so we choose it to demonstrate the generalization and robustness of GDP. We use ResNet-50 as backbone, and follow the same training settings as public implementation~\footnote{https://github.com/VainF/DeepLabV3Plus-Pytorch}. The output stride (OS) for training and validation is 16. Table~\ref{tab:voc} shows the mIoU of Pascal VOC segmentation task, and several visual results are shown in supplementary materials. For the sub-net with 38.4\% FLOPs, we are surprised to find that the branch with atrous-rate 18 in the ASPP module is totally pruned, the branch with atrous-rate 12 only 6 channels left, however, gaining slightly better performance.

Furthermore, we also deploy GDP method in \textbf{Style Transfer} task, based on official PyTorch implementation of~\cite{johnson2016perceptual}, achieving great performance on pruned models with large portion of FLOPs savings. Visual results of pruned models are shown in supplementary materials.

\begin{table}
	\centering
	\begin{center}
	\setlength{\tabcolsep}{1.2mm}{
	\begin{tabular}{cccccc}
		\toprule
		FLOPs(\%)   &   38.4          & 23.3  & 12.1 & 5.2 &Baseline\\
		\midrule
        mIOU            & \textbf{0.779}  &	0.767 &  0.721& 0.666    & 0.774 	\\
		\bottomrule
	\end{tabular}
	}
	\end{center}	
	\caption{The mIoU on Pascal VOC 2012 validation set with DeepLabv3+ using ResNet-50 as backbone. FLOPs here means the remaining FLOPs ratio of the sub-net. We can see that GDP can obtain a sub-net containing only 38.4\% FLOPs without performance drop. More visual results are in supplementary materials. }
	\label{tab:voc}
\end{table}

\subsection{Discussion}
\begin{figure*}[ht!]
\centering
\subfigure[Initialization]{
\begin{minipage}[t]{0.13\textwidth}
\centering
\includegraphics[width = 0.98\textwidth]{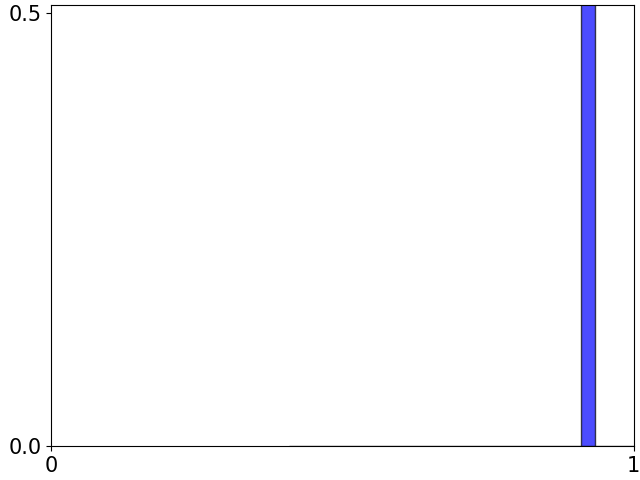}
\end{minipage}
}\subfigure[epoch 0]{
\begin{minipage}[t]{0.13\textwidth}
\centering
\includegraphics[width = 0.98\textwidth]{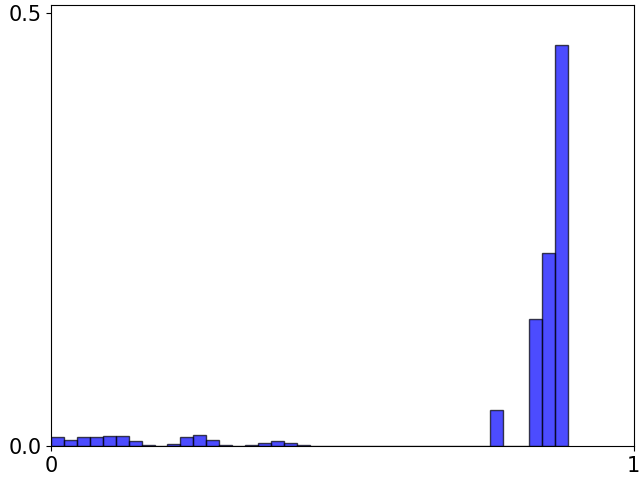}
\end{minipage}
}\subfigure[epoch 10]{
\begin{minipage}[t]{0.13\textwidth}
\centering
\includegraphics[width = 0.98\textwidth]{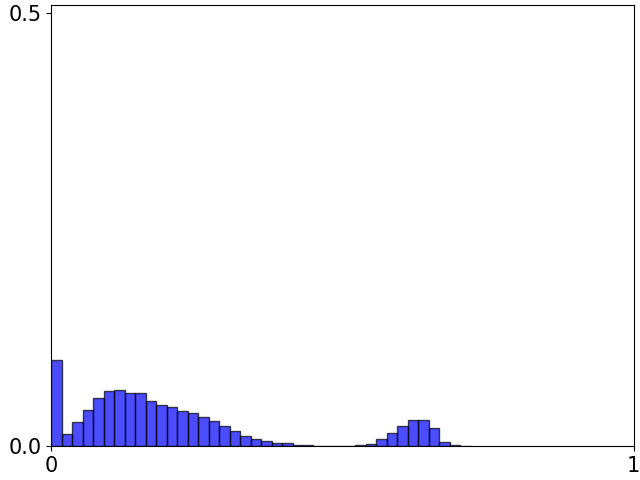}
\end{minipage}
}\subfigure[epoch 20]{
\begin{minipage}[t]{0.13\textwidth}
\centering
\includegraphics[width = 0.98\textwidth]{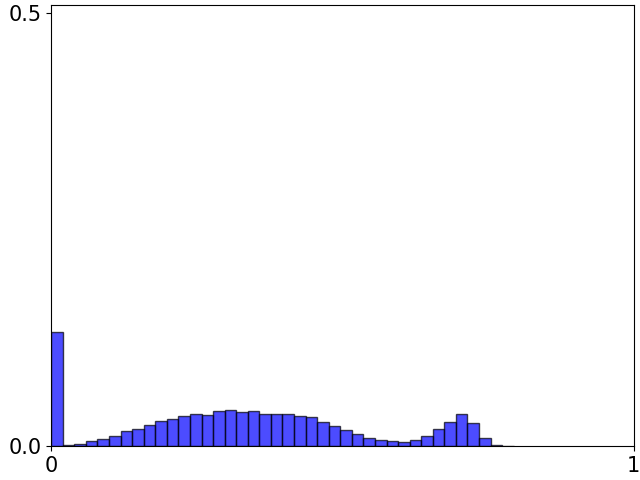}
\end{minipage}
}\subfigure[epoch 40]{
\begin{minipage}[t]{0.13\textwidth}
\centering
\includegraphics[width = 0.98\textwidth]{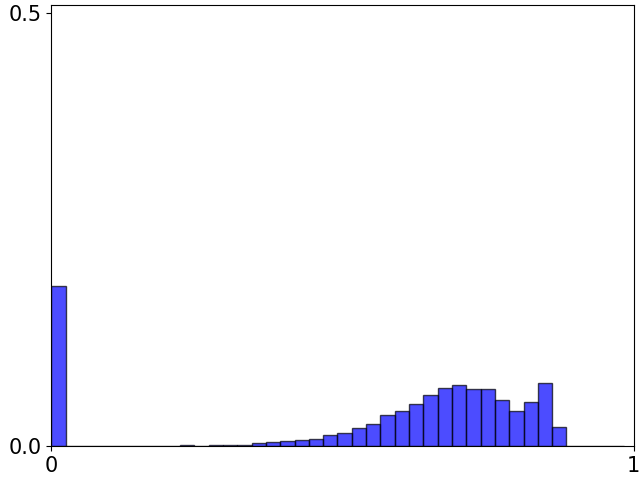}
\end{minipage}
}\subfigure[epoch 70]{
\begin{minipage}[t]{0.13\textwidth}
\centering
\includegraphics[width = 0.98\textwidth]{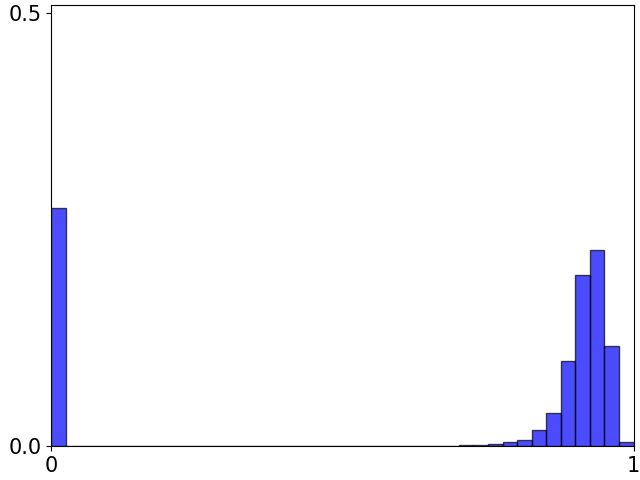}
\end{minipage}
}\subfigure[epoch 139]{
\begin{minipage}[t]{0.13\textwidth}
\centering
\includegraphics[width = 0.98\textwidth]{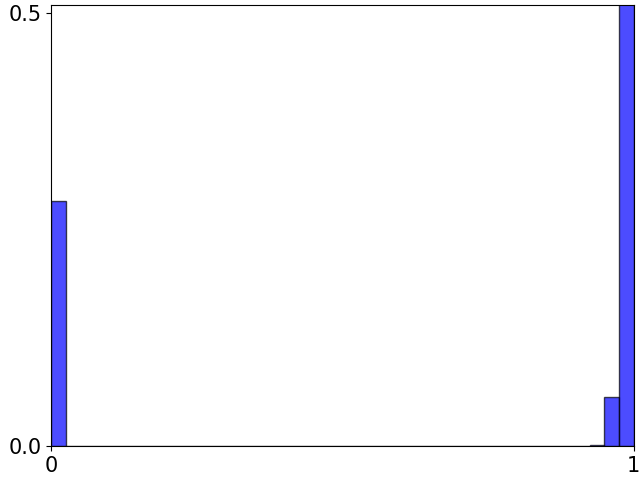}
\end{minipage}
} \\
\caption{The density histogram of gates as the training goes on. All the X-axis represents the gate values in $[0, 1]$, Y-axis represents the frequency of gates (we truncate it to 0.5 for better visualization). At beginning, the $\alpha$ and $\epsilon$ are initialized 1.0 and 0.1 respectively, so all the gates are $\frac{1}{1.1} \approx 0.909$ (see Function \eqref{eq:smoothed l0}) at the beginning and gates are distributed smoothly between 0 and 1. After some epochs, the gates becomes more and more polarized. Finally, some gates become exact zeros while others close to ones. The figure is based on experiment conducted on Imagenet with MobileNet-V2, the final remaining FLOPs ratio is 0.45.}
\label{fig: midu}
\end{figure*}

\paragraph{Polarization} 
\rev{Figure \ref{fig: midu} shows density distributions of gate values during the training process.
We can see that the density distributions of gates change smoothly and the number of zero-gates becomes stable after epoch 70. The polarization depends on the value of $\epsilon$. A larger $\epsilon$ results in a smoother gate with less polarization; a smaller $\epsilon$ makes ~\eqref{eq:smoothed l0} closer to $L_0$ norm but with less numerical stability. An ablation study for the effects of $\epsilon$ is attached in supplementary materials. Our method is robust to different values of $\epsilon$.}

\begin{figure}[h!]
  \includegraphics[width = 0.24\textwidth]{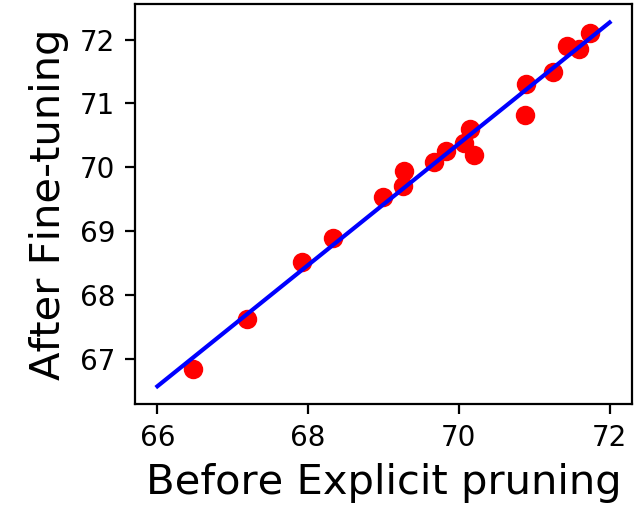}
  
  \vspace*{\dimexpr-\parskip-8.5\baselineskip}
  \parshape 10 
    .26\textwidth 0.23\textwidth 
    .26\textwidth 0.23\textwidth 
    .26\textwidth 0.23\textwidth
    .26\textwidth 0.23\textwidth
    .26\textwidth 0.23\textwidth
    .26\textwidth 0.23\textwidth
    .26\textwidth 0.23\textwidth
    .26\textwidth 0.23\textwidth
    .26\textwidth 0.23\textwidth
    0pt 0.5\textwidth 
  \makeatletter
  \refstepcounter\@captype
  \addcontentsline{\csname ext@\@captype\endcsname}{\@captype}
    {\protect\numberline{\csname the\@captype\endcsname}{ToC entry}}%
  \csname fnum@\@captype\endcsname: 
  \makeatother
  \small {The relationship between Top-1 accuracy of super-net before explicitly pruning and that of the corresponding sub-net after fine-tuning. The correlation coefficient is 0.993. We can see that the accuracy of super-net effectively reflects that of the final sub-net. This experiments are conducted on ImageNet with MobileNet-V1.}
  \label{fig:consistency}
\end{figure}

\begin{figure}[h!]
\centering
\subfigure[train/val loss ]{
\begin{minipage}[t]{0.23\textwidth}
\centering
\includegraphics[width = 1.0\textwidth]{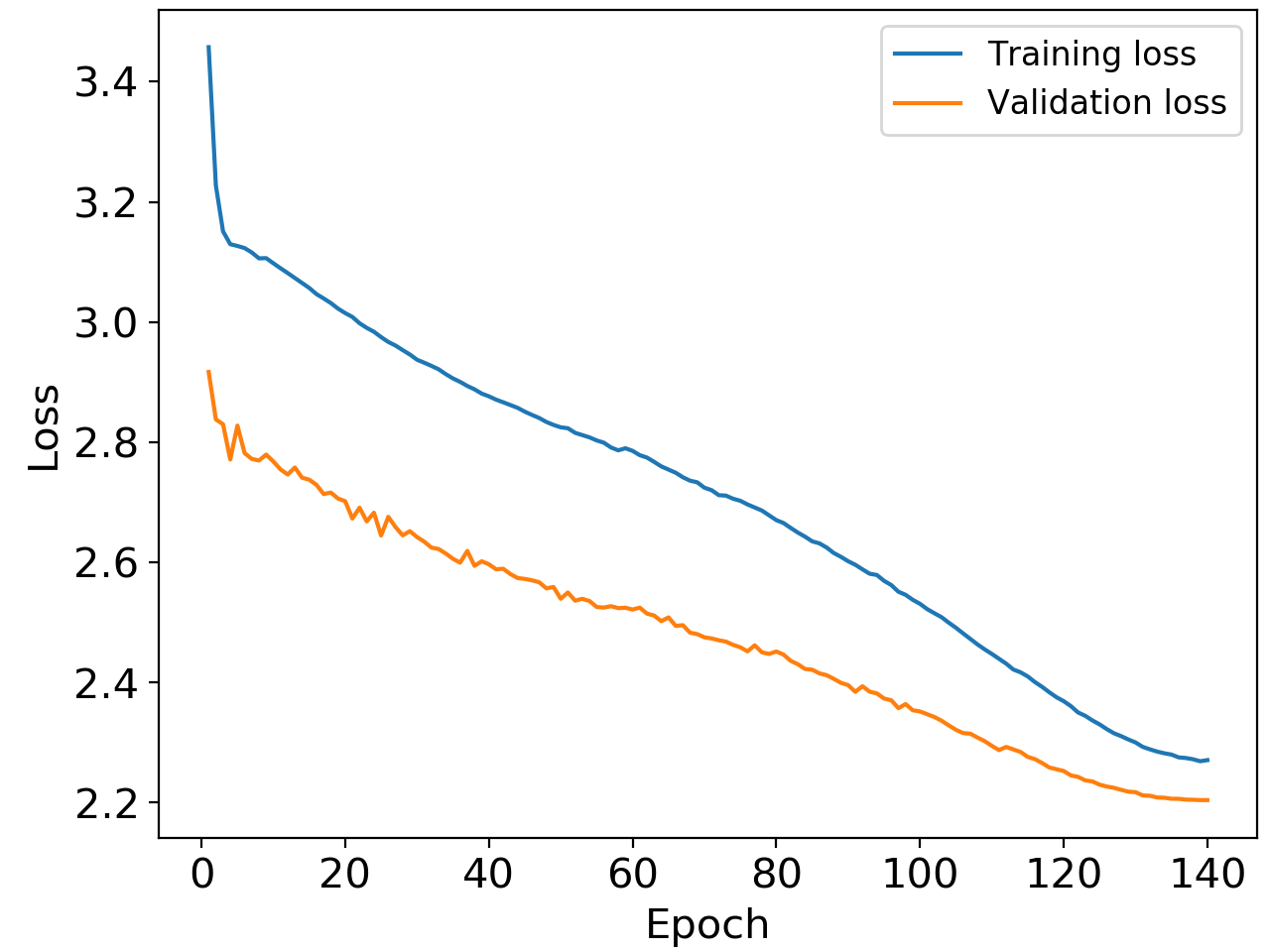}
\end{minipage}
}\subfigure[Remaining FLOPs ratio]{
\begin{minipage}[t]{0.23\textwidth}
\centering
\includegraphics[width = 1.0\textwidth]{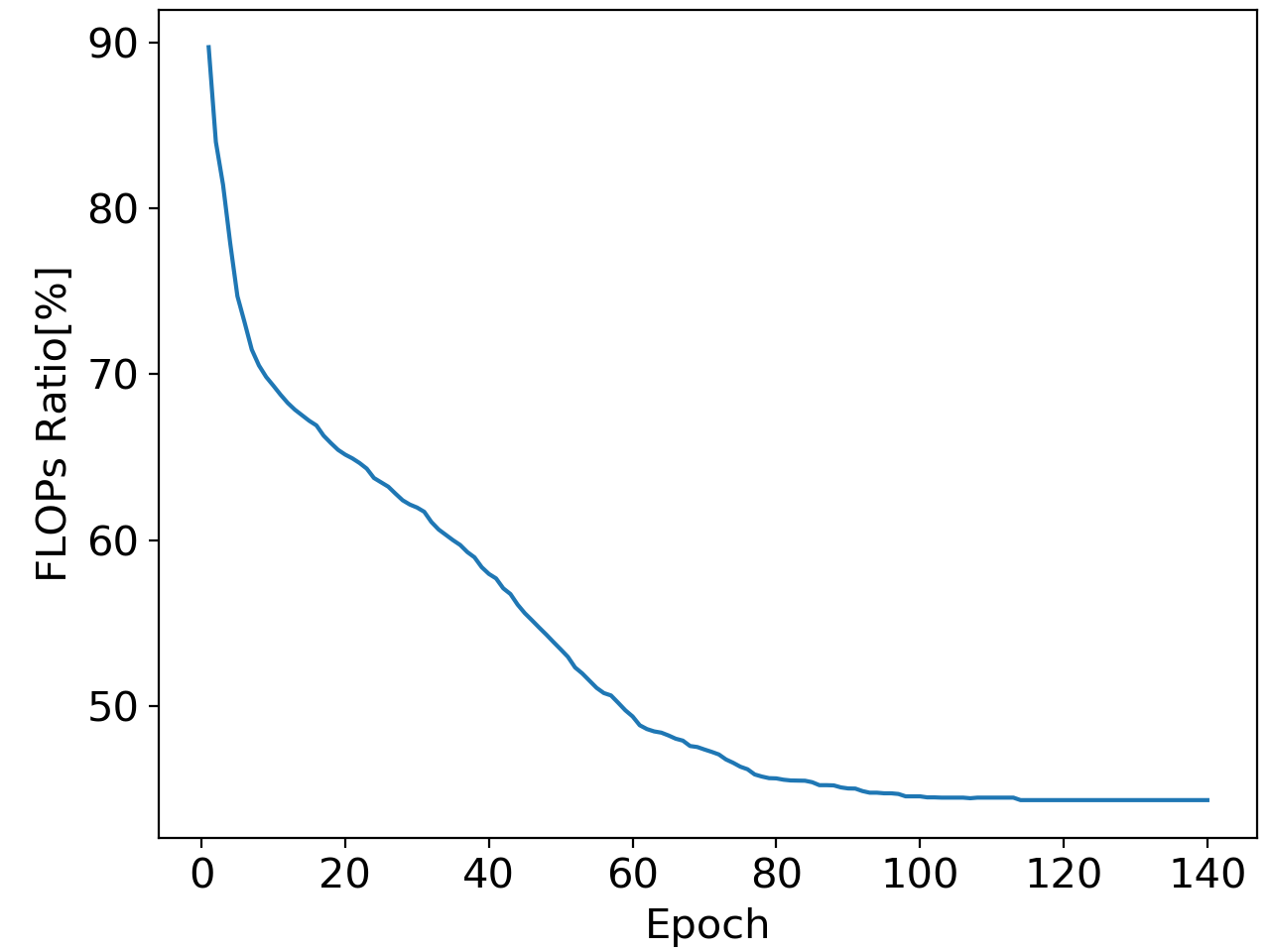}
\end{minipage}
}
\caption{The stability of GDP. We can clearly see that the
training/validation loss, 
and the remaining FLOPs ratio is changing stably as training goes on. This figure is plotted on ImageNet with MobileNet-V2. As we decay $\epsilon$ in Function \eqref{eq:smoothed l0} after each epoch, the BN parameters are re-calculated for plotting validation loss. All the loss is averaged within each epoch, while the FLOPs ratio is calculated at the end of each epoch.}
\label{fig: stability}
\end{figure}

\paragraph{Consistency} We do not need to sample sub-net or calculate expectation when training. Figure~\ref{fig: stability} shows the training stability of GDP. When training terminates, there is only one candidate sub-net left. Different value of $\lambda$ in Function \eqref{eq:objective} result in sub-net with different FLOPs and performance. We want the performance of the super-net to reflect the performance of the final sub-net as much as possible. We set multiple hyper-parameters to run the GDP compression training. 
Figure~\ref{fig:consistency} reflects the relationship between performance of super-net before explicitly pruning and the sub-net after fine-tuning. 


\begin{figure}[ht!]
  \includegraphics[width = 0.26\textwidth]{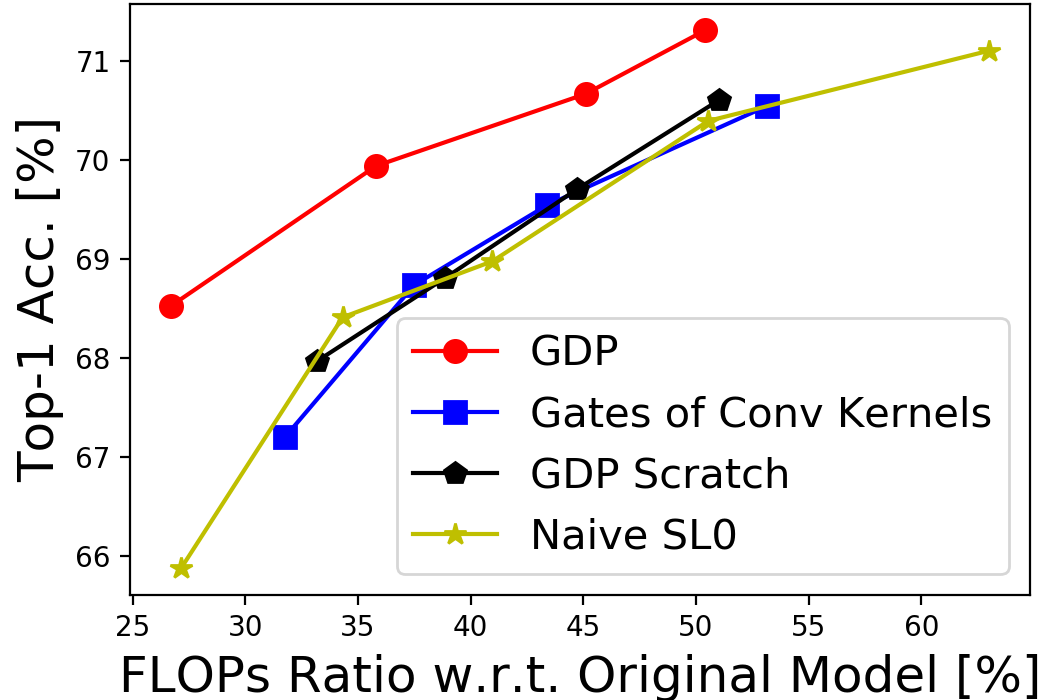}
  
  \vspace*{\dimexpr-\parskip-7.4\baselineskip}
  \parshape 10 
    .27\textwidth 0.21\textwidth 
    .27\textwidth 0.21\textwidth 
    .27\textwidth 0.21\textwidth
    .27\textwidth 0.21\textwidth
    .27\textwidth 0.21\textwidth
    .27\textwidth 0.21\textwidth
    .27\textwidth 0.21\textwidth
    .27\textwidth 0.21\textwidth
    .27\textwidth 0.21\textwidth
    0pt 0.5\textwidth 
  \makeatletter
  \refstepcounter\@captype
  \addcontentsline{\csname ext@\@captype\endcsname}{\@captype}
    {\protect\numberline{\csname the\@captype\endcsname}{ToC entry}}%
  \csname fnum@\@captype\endcsname: 
  \makeatother
  \small {Ablation studies on Imagenet with MobileNet-V1. ``Gates of Conv kernels" and ``Naive SL0" represent Function~\eqref{eq:not alpha} and ~\eqref{eq:original SL0} respectively. ``GDP Scratch" means pruning from scratch by GDP.}
  \label{fig:g(w)}
\end{figure}

\subsection{Ablation studies}
\paragraph{Not introducing $\alpha$} There is an alternative solution that instead of introducing the learnable parameters $\alpha$, we can also use the existing convolution kernels weights as $x$ in Equation \eqref{eq:smoothed l0}. The objective function is then becomes:
\begin{equation}
    \min_{w}\mathcal{F}(w) = \mathcal{L}(w;g(w))+\lambda \mathcal{R}(w)
    \label{eq:not alpha}
\end{equation}
For the $l$-th layer convolution kernel $W_l \in \mathbb{R}^{c\times d \times r \times r}$, we can let $x_l^{(i)} = \|W_{l[:,i,:,:]}\|_2$ where $i=1,2,...,d$, is the indication of input channels, and 
substitute $x_l$ defined here for $\alpha$ in Equation~\eqref{eq:resource-rewritten}. In this case, we initialize $\epsilon_l^{(i)}$ in every gate to one-tenth of the corresponding $\|W_{l[:,i,:,:]}\|_2$, so the initial value of all gates is $\frac{1}{1.1} \approx 0.909$, the same to when $\alpha$ is introduced. The performance is plotted in Figure \ref{fig:g(w)}, labeled with ``Gates of Conv Kernels". We guess that the relatively poor performance of using $\|W_{l[:,i,:,:]}\|_2$ may be caused by dual roles of both feature extraction and gate controlling, which has the opposite thoughts from ~\cite{salimans2016weight,wang2016dueling}, or caused by the occasional instability of the norm of $w$ when $\epsilon$ is small enough.

\paragraph{$g$ as regularization} $g(x)$ was originally used as regularization term added to the original objective function of smooth L0 optimization~\cite{xiang2019new}. Follow this idea, we can also naively define our objective function as bellow:
\begin{equation}
    \min_{w}\mathcal{F}(w) = \mathcal{L}(w)+\lambda \mathcal{R}(g(w))
    \label{eq:original SL0}
\end{equation}
The reason we did not take this approach is similar to above. The norm of $w$ is occasionally unstable when training, maybe caused by $g(w)$. However, if we introduce $\alpha$ as in Equation~\eqref{eq:objective}, it is not $\alpha$ itself but $g(\alpha)$ to participate in the network forward. $g(\alpha)$ is stably taking values in range $[0,1]$. The performance is shown in Figure~\ref{fig:g(w)}, labeled with ``Naive SL0".

\paragraph{Pruning from Scratch} 
Figure~\ref{fig:g(w)} also shows some results by pruning from scratch (labeled with ``GDP Scratch") \rev{rather than from a well-trained model}. \rev{With the same hyper-parameters, the pruning speed of ``GDP Scratch" became particularly faster and a more compressed sub-net was obtained. This relatively poor performance might result from that some important channels have no resistance to being pruned before they get learned well enough.}

\section{Numerical Result}
We put the numerical result in Table~\ref{mobileVe} and Table~\ref{mobileV2} for better comparison with future work, although it is already attached in supplementary materials.

\begin{table}[!h]
	\centering
	\begin{center}
	\setlength{\tabcolsep}{1mm}{
 	\begin{tabular}{ccccccc}
		\toprule
		FLOPs(\%) & 26.7 & 35.8 & 45.1 & 50.4 & 58.3 & 71.4\\
		\midrule
        MV1(top1)    & 68.5 & 69.9 & 70.7 & 71.3 & 71.9 & 72.4\\
		\bottomrule
	\end{tabular}

	 \begin{tabular}{ccccccc}
		\toprule
		FLOPs(\%) & 18.9 & 25.4 & 30.4 & 34.4 & 38.7 & 49.0\\
		\midrule
        Res50(top1) & 74.1 & 75.1 & 75.6 & 76.1 & 76.4 & 77.1\\
		\bottomrule
	\end{tabular}

	}
	\end{center}	
	\caption{The compression ratios and corresponding Top 1 accuracy of MobileNet\_v1(MV1), and ResNet50(Res50)}
	\label{mobileVe}
\end{table}

\begin{table*}[!h]
	\centering
	\begin{center}
	\setlength{\tabcolsep}{1mm}{

	 \begin{tabular}{cccccccccccc}
		\toprule
		FLOPs(\%) & 13.0 & 16.5 & 22.0 & 26.4 & 30.4 & 36.1 & 43.8 & 52.1 & 62.4 & 77.6 & 96.1\\
		\midrule
        MV2(top1) & 60.3 & 62.6 & 65.4 & 66.8 & 67.8 & 68.9 & 70.2 & 70.9 & 71.8 & 72.5 & 73.1\\
		\bottomrule
	\end{tabular}

	}
	\end{center}	
	\caption{The compression ratios and corresponding Top 1 accuracy of MobileNet\_v2(MV2)}
	\label{mobileV2}
\end{table*}

\section{Ablation study for $\epsilon$}
\begin{figure}
\centering
\includegraphics[width = 0.41\textwidth]{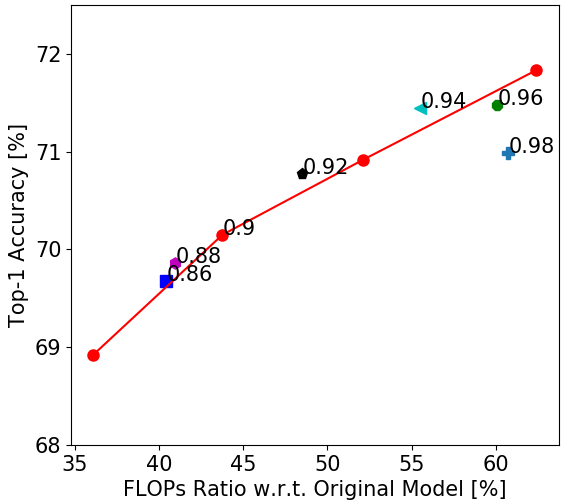}
\caption{This figure shows different decaying rates of $\epsilon$ and their corresponding compression ratios and Top-1 accuracy. The experiment is conducted on ImageNet with MobileNet-V2. All the discrete dots have all the same hyper-parameters except the decaying rate, and the attached numerical values represent the decaying rate for each epoch. The red dashed polyline is from the main body of the paper only for reference. All the initial values of $\epsilon$ is 0.1. We can clearly see that GDP is somewhat robust to the decaying rate of $\epsilon$ ranging from 0.86 to 0.96, which results in the final $\epsilon$ ranging from $0.1*0.86^{140}\approx 6e^{-11}$ to $0.1*0.96^{140}\approx 3e^{-4}$.}
\label{fig: epsilon_decay}
\end{figure}

The smooth L0 function is defined as:
\begin{equation}
    g_{\epsilon}(x) = \frac{x^2}{x^2 + \epsilon}
    \label{eq:smoothed l0}
\end{equation}
The polarization depends on the value of $\epsilon$. A larger $\epsilon$ results in a smoother gate with less polarization; a smaller $\epsilon$ makes gate function closer to $L_0$ norm but with less numerical stability. Fig.~\ref{fig: epsilon_decay} shows different decaying rates of $\epsilon$ with their corresponding compression ratios and Top-1 accuracy. We can see that GDP is robust to the final value of $\epsilon$ ranging from $0.1*0.86^{140}\approx 6e^{-11}$ to $0.1*0.96^{140}\approx 3e^{-4}$, which is a really big range.

\begin{figure}
\centering
\includegraphics[width = 0.41\textwidth]{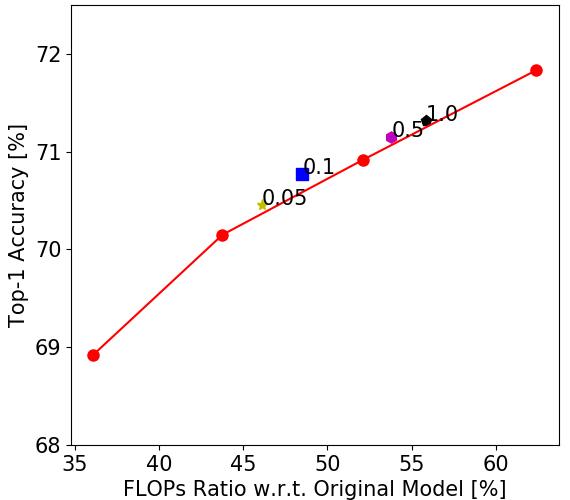}
\caption{This figure shows different initial values of $\epsilon$ and their corresponding compression ratios and Top-1 accuracy. The experiment is conducted on ImageNet with MobileNet-V2. All the discrete dots have all the same hyper-parameters except the initial value of $\epsilon$, and the attached numerical values represent the initial value. The red dashed polyline is from the main body of the paper only for reference. All the decaying rate of $\epsilon$ is 0.92 for each epoch. We can clearly see that GDP is somewhat robust to the initial value of $\epsilon$.}
\label{fig: epsilon_init}
\end{figure}

Besides the decaying rate, we also study the effect of initial value of $\epsilon$, as shown in Fig.~\ref{fig: epsilon_init}. We can also see the robustness of GDP with respect to different initial values.

\section{Visual results for style transfer and semantic segmentation}
\begin{figure*}
\centering
\includegraphics[width =\textwidth]{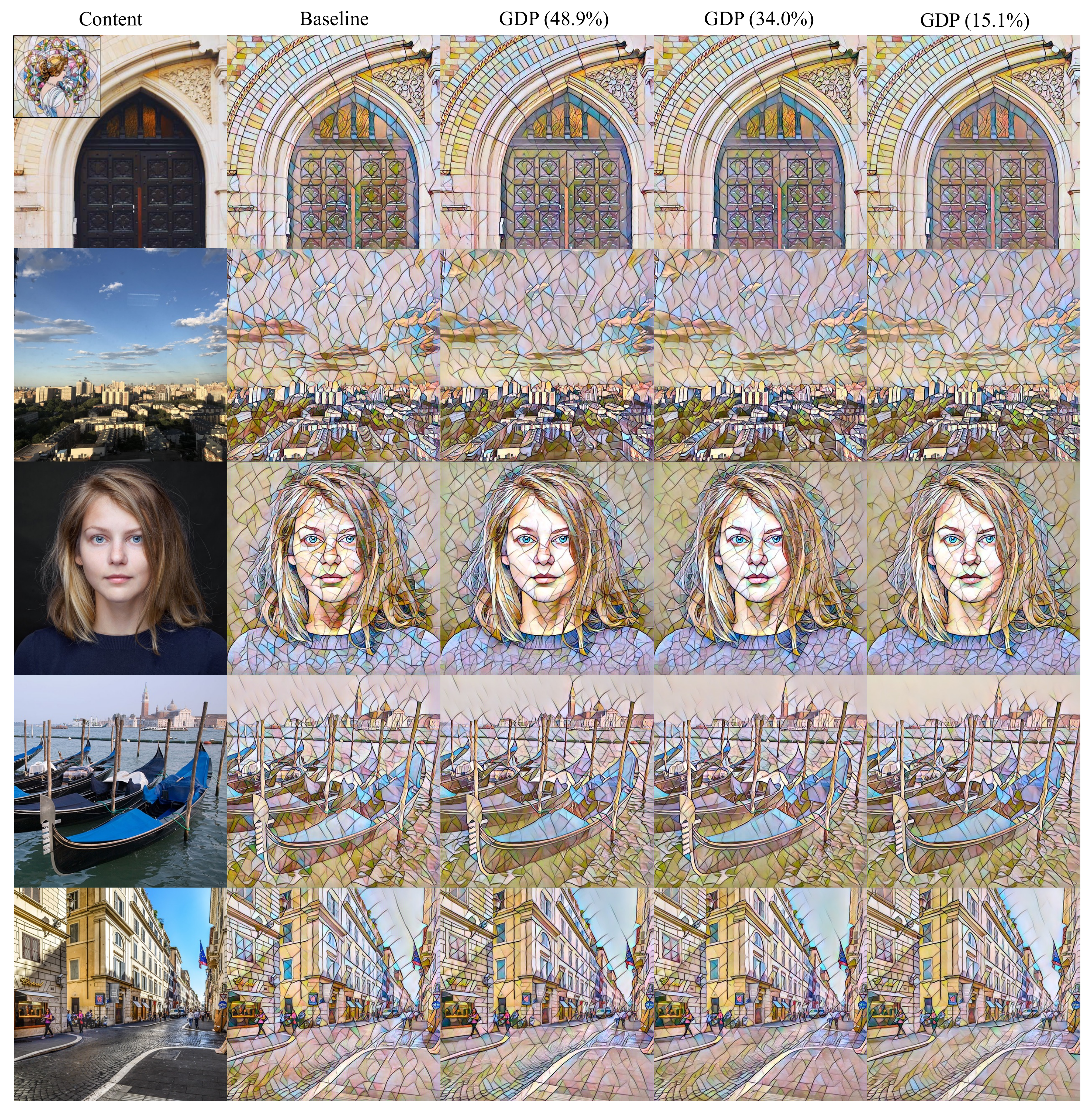}
\caption{Visual comparisons between baseline and our GDP pruned models. GDP (48.9\%) means that the remaining flops ratio of the sub-net pruned by GDP is 48.9\%}
\label{fig: fns-1}
\end{figure*}

\begin{figure*}
\centering
\includegraphics[width =\textwidth]{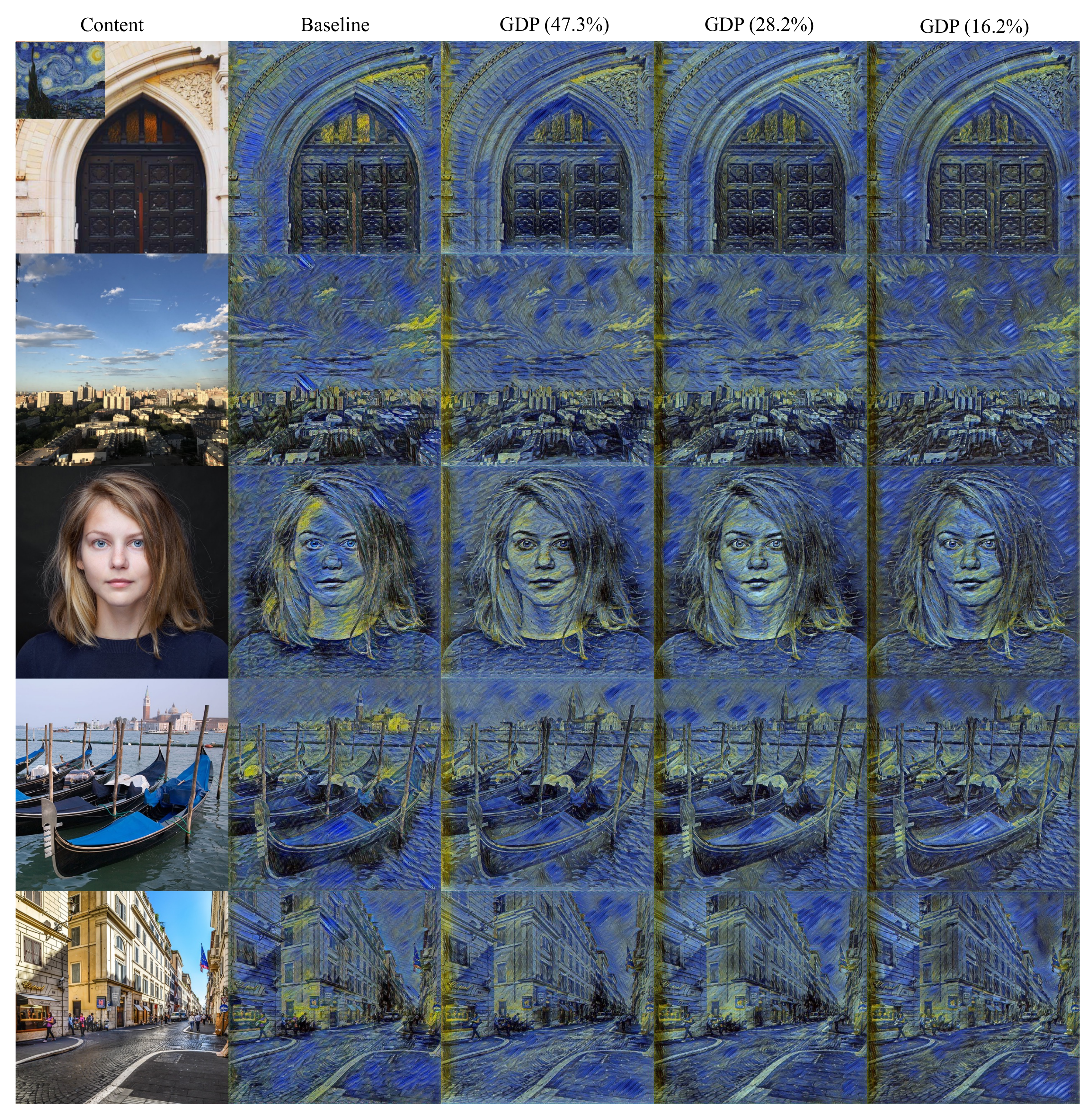}
\caption{Visual comparisons between baseline and our GDP pruned models on style transfer. GDP (47.3\%) means that the remaining flops ratio of the sub-net pruned by GDP is 47.3\%}
\label{fig: fns-2}
\end{figure*}

\begin{figure*}
\centering
\includegraphics[width=0.8\textwidth]{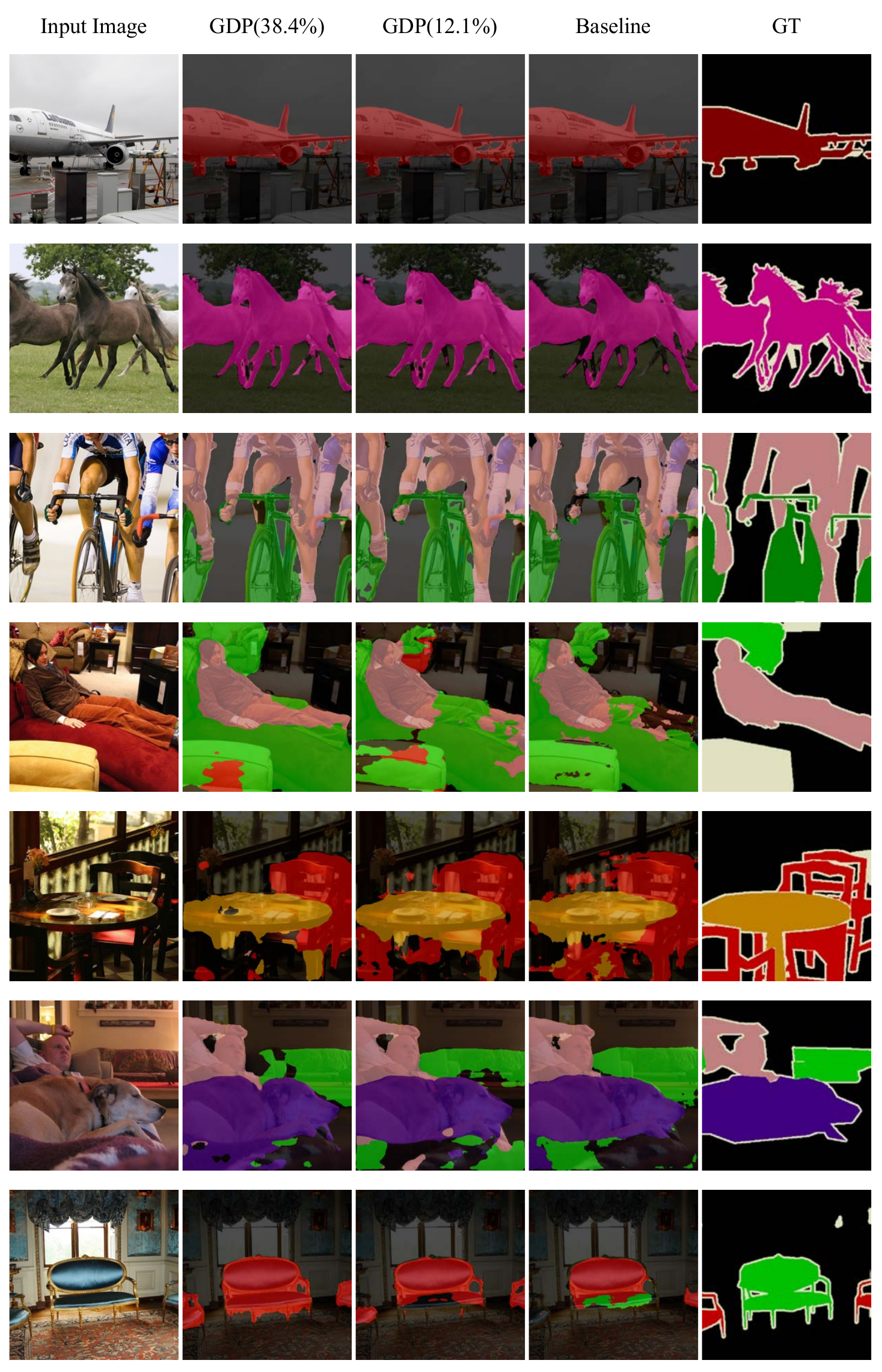}
\caption{Visual comparisons between baseline and our GDP pruned models for Pascal VOC with DeepLabv3+. We can find our pruned model can even achieve better boundary quality.}
\label{fig: seg_results}
\end{figure*}

For style transfer, it can be seen that our pruned model can achieve similar results without hurting perceptual quality, as shown in Fig.~\ref{fig: fns-1} and Fig.~\ref{fig: fns-2}. Also, our pruned model can even achieve better boundary quality compared with and semantic segmentation baseline, as shown in Fig.~\ref{fig: seg_results}

\section{Conclusion}
In this paper, we proposed a novel differentiable polarized gate module in network pruning task and obtain state-of-the-arts performance on various benchmark experiments. Comparing with other optimization-based pruning methods, the advantages of our method are five folds: simpler to implementation; fewer hyper-parameters to tune; obtain good sub-nets easily; take less time; not rely on any specific network structure. 
The disadvantage is that the model pruning ratio cannot be specified in advance, which now need to manually adjust the value of $\lambda$ to balance the accuracy and resource consumption of network. 
We leave this problem in future works. 
The gates with differentiable polarization proposed in this paper can also be easily extended into neural architecture search (NAS) tasks.

\clearpage
\clearpage

{\small
\bibliographystyle{ieee_fullname}
\bibliography{egbib}
}

\end{document}